\documentclass[fleqn,10pt]{wlscirep}
\title{The ``handedness'' of language: Directional symmetry breaking
of sign usage in words}

\author[1,2]{Md Izhar Ashraf}
\author[1,3,*]{Sitabhra Sinha}
\affil[1]{The Institute of Mathematical Sciences, CIT Campus,
Taramani, Chennai 600113, India.}
\affil[2]{B. S. Abdur Rahman University, Seethakathi Estate,
Vandalur, Chennai 600048, India.}
\affil[3]{National Institute of Advanced Study, Indian Institute of
Science Campus, Bangalore 560012, India.}

\affil[*]{sitabhra@imsc.res.in}

\begin{abstract}
Language, which allows complex ideas to be
communicated through
symbolic sequences, is
a characteristic feature of our species and manifested in a multitude
of forms.
Using large written corpora for many different languages and scripts, 
we show that the occurrence probability distributions of signs at the
left and right ends of words have a distinct heterogeneous nature.
Characterizing this asymmetry using quantitative inequality 
measures, viz. 
information entropy and the Gini index,
we show that the beginning of a word is less restrictive in sign usage
than the end.  
This property is not simply attributable to the use of
common affixes as it is seen even when only word roots are considered.
We use the existence of this asymmetry to infer the direction of
writing in undeciphered inscriptions that agrees with the
archaeological evidence. Unlike traditional investigations of
phonotactic constraints which focus on language-specific patterns, our
study reveals a property valid across languages and writing systems.
As both language and writing are unique aspects of our species, this
universal signature may reflect an innate feature of the human
cognitive phenomenon.
\end{abstract}
\begin{document}

\maketitle

\thispagestyle{empty}

\section*{Introduction}
Language - and by extension, writing - distinguishes humans from all
other species~\cite{Deacon1997}. The ability to communicate 
complex information across both space and time have enabled society
and civilization to emerge~\cite{Dunbar1996}. The recent availability
of publicly accessible ``big data'', such as 
the large digitized corpus on the {\em Google Books}
website, has revolutionized the quantitative analysis of socio-cultural
phenomena and led to new empirical discoveries~\cite{Michel2011,Petersen2012,Dodds2015}. 
Language in its written form is represented as symbolic sequences that convey
information. 
Statistical analysis of such sequences have led to the identification of several
quantitative properties that hold across many human languages. For example,
one of the best known empirical
regularities associated with language is the scaling behavior - referred 
to as Zipf's law - that quantifies how some words occur far more frequently than others~\cite{Zipf1932}.
Several possible theoretical explanations of the phenomenon have been
proposed~\cite{Mitzenmacher2003,Cancho2003}. Words are themselves
composed of  
signs corresponding to letters, syllabograms or logograms depending on the writing system.
It has long been known that the different
signs, e.g., letters, also occur with characteristic frequencies - a
fact that has been used by cryptographers over the ages to break
simple substitution ciphers. This was 
illustrated dramatically in fiction
by Poe ({\em The Gold-Bug}) and Conan Doyle ({\em The
Adventure of the Dancing Men}). 
For English, the phrase ``ETAOIN SHRDLU" has 
often been used as a mnemonic for recalling
the approximate order of the most commonly occurring letters in typical texts.
However, a cursory glance through an English dictionary (or
encyclopedia) to ascertain,
for each letter of the alphabet, the number of pages that are required
to list all the words (or entries) that begin with
that letter, will alert one to a strong deviation
from what is naively expected from the frequency distribution of
letters. For instance, one of the letters having the largest number of entries
in a dictionary
is `c' which does not even appear among the most frequently used letters
in English as per the phrase above.
This apparent anomaly arises from the fact that the letter `c' has 
a much higher probability of occurring (relative to other letters)
at the beginning of an English word - possibly a result of the
specific orthography of English, where it can appear as the initial
letter of the words
{\em china}, {\em can}, {\em cent}, etc., in all of which it is pronounced
differently - but does not occur so frequently at other
positions. While it is rarer to come across situations where words are
arranged according to their last character, it is possible to
ask whether the frequency distribution of the letters that occur at the 
end of a word will
similarly show a distinct character.
For example, had we arranged the entries of an English dictionary in the order of the 
{\em last} letter of each word, we would see that this would result in the number of entries 
corresponding to each letter showing a far more unequal distribution than seen in a
conventional dictionary. In other words, the last position in an English word is
usually occupied by one of very few letters, suggesting that the final letter is much
more tightly constrained than the initial letter. We show that this is not just true for
English but holds in at least 23 other languages, including those which use writing systems not based on letters (alphabets and abjads), but instead on signs representing syllables or logograms. 
We have also applied this property of asymmetry in sign usage patterns between the beginning 
and end of a sequence to an undeciphered corpus of inscriptions, showing that we can determine the direction in which the sequences were written (left to right or right to left).

\begin{figure*}[tbp]
\begin{center}
\includegraphics[width=0.99\linewidth]{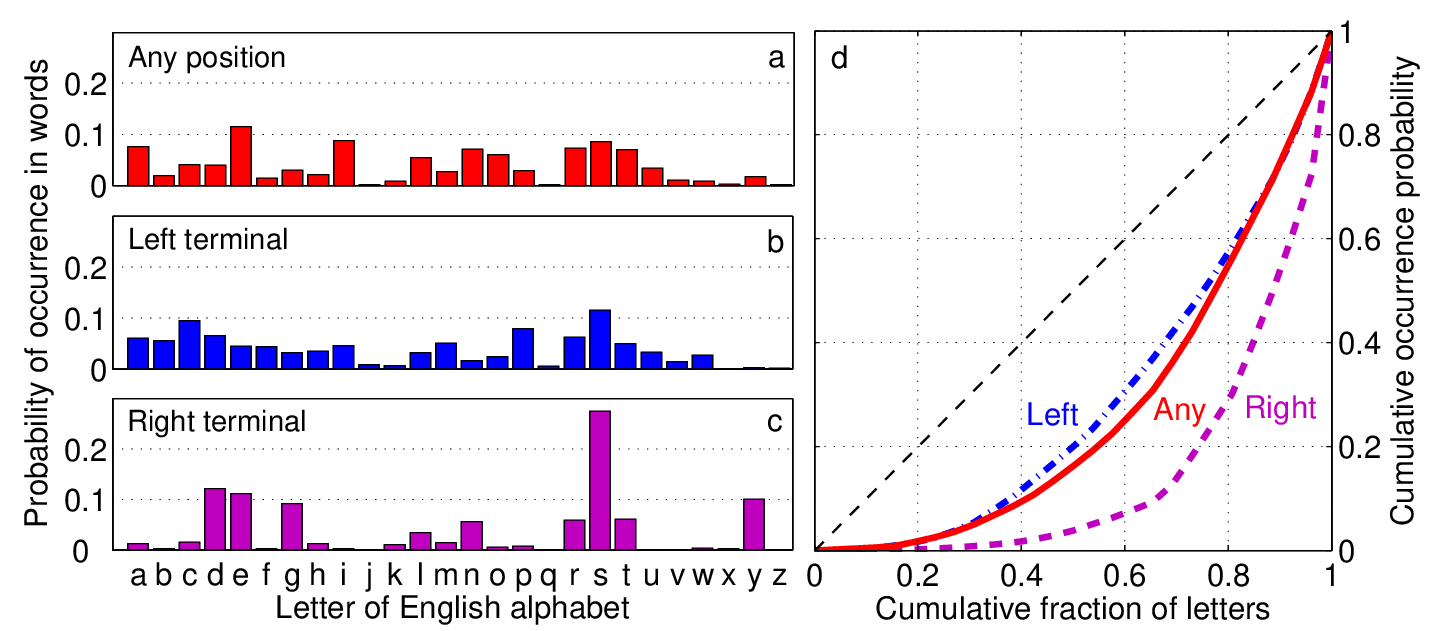}
\end{center}
\caption{
{\bf Unequal representation of letters (1-grams) occurring at
different positions in words in written English.} The probability of
occurrence of the 26 letters of the English alphabet in the
\textit{Mieliestronk} corpus comprising about 58000 unique words of
the English language (see Methods for details),  at (a) any position,
(b) left terminal position (i.e., in the beginning) and (c) right
terminal position (i.e., at the end) of a word. The distribution shows
more heterogeneity in the letter occurrence
probabilities for (c), indicating that only a few letters occur
with high frequency at the right terminal position of a word, compared
to a relatively more egalitarian frequency of occurrence of letters in
the left terminal position (b). This difference is illustrated in the
Lorenz curve (d) comparing the cumulative distribution function for
the occurrence probability of the different letters in any (solid
curve), left terminal (dash-dotted curve) and right terminal position
(dashed curve) of a word. The thin broken diagonal line
(line of perfect equality)
corresponds to
a perfectly uniform distribution, deviation from which indicates the
extent of heterogeneity of letter occurrence probability distributions
- measured by the Gini index which is the ratio of the
  area between the line of perfect
equality and the observed Lorenz curve, and, the area between the lines
of perfect equality and of perfect inequality (viz., the horizontal
line).}
\label{fig1}
\end{figure*}

Fig~\ref{fig1}~(a-c) shows the occurrence probability distribution of
the 26 letters of the Latin alphabet at the initial and final
positions of English words reconstructed from a large database, and
compares it with their probabilities of occurring anywhere in a text.
We note immediately that letters may differ greatly in terms of their
occurrence probability depending on the position - but most
importantly, the distribution for the right terminal character (the last
letter) in an English word appears to be much more heterogeneous than
the one corresponding to the left terminal character (the first
letter).
In other words, the choice of letters that can occur as the final
character of a word is more restrictive, i.e., the occurrence
probabilities are more unequal, with very few accounting for
the right terminal position for a major fraction of the words,
compared to their position-independent probabilities. In contrast, the
probability distribution of letters that occur as the initial
character is more egalitarian (in comparison to the
final one), implying a somewhat higher degree of
freedom of choice at the left terminal position.
To ensure that this left-right asymmetry in sign usage distributions -
suggesting a ``handedness'' of words in terms of the letter frequency
distributions at their terminal positions - is not an artifact of the
corpus one is using, we performed the same analysis with
{\em Google Books Ngram} data, focusing on words that occur with a
frequency of more than $10^5$ in the corpus digitized by
Google (see Supporting Information for details).
As seen from Fig~\ref{figs1}, the qualitative features are similar to
that observed in Fig~\ref{fig1}, indicating the
robustness of the observed left-right asymmetry of sign occurrence
probability patterns in words.

In this paper we will argue that this directional asymmetry is not 
just a feature
of a particular language but appears to be universal, holding
across different languages and writing systems. Regardless of whether the
signs we are considering represent letters (for alphabetic scripts
like English), syllabograms (for syllabic scripts such as Japanese
Kana) or logograms (for logographic scripts like Chinese or
logo-syllabic ones like Sumerian cuneiform), the distribution of the
signs that begin a word shows relatively less heterogeneity than that for
the ones that occur at its end. We have used measures of
inequality (viz. Gini index and information entropy) to
quantitatively assess the degree of asymmetry in the sign occurrence 
distributions for different linguistic corpora.
The difference in the two distributions also indicate the differential
information contents of the initial and final characters - and links
our result to the statistical and information-theoretic analysis of 
language~\cite{Mumford2010}. This approach was pioneered by Shannon
who used
the concept of predictability, i.e., the constraints imposed on a
letter by those that have preceded it, to estimate the
bounds for the entropy (the amount of information per letter)
and redundancy in English~\cite{Shannon1948,Shannon1951}. 
%
Considering the
consequences of the most
prominent structural patterns of texts - viz., the clustering of
letters into words - Sch\"{u}rmann and Grassberger subsequently 
showed that the 
the average entropies of letters located inside a word are much
smaller than that of the letters at the 
beginning~\cite{Schurmann1996,Schurmann1996b}. However, this is true
even if one reverses the word - so that terminal letters of words (whether
initial or final) have less predictability than those in other
positions. Here we ask the relatively simpler question of whether 
the statistical properties of the left and right terminal 
characters are different and find a surprising non-trivial asymmetry
in the heterogeneity of the respective distributions.
Analysis of correlation between sign occurrences in written texts have 
traditionally
focused on the phonotactic constraints of specific languages, e.g., 
determining the consonants or consonant clusters that are allowed
to occur before and after a vowel in any syllable of a given language.
While there is considerable variation between different languages as
regards the possible arrangements in which consonants and vowels 
can be combined to make meaningful words, here we show the existence
of general patterns that hold across many different language
families. 

\section*{Results}
\label{sec2}
In order to quantify the heterogeneity in sign usage
distribution at 
the beginning and at the end of a word, we have
used the Gini index or coefficient~\cite{Gini1912}.
It measures dispersion in the distribution of a
quantity and is widely used in the socio-economic literature to quantify
the degree of 
inequality, e.g., in the distribution of income of individuals or
households~\cite{Sinha2011}. The value of the Gini 
index $G$ (see Eq.~\ref{ginieqn} in Materials and
Methods) expresses the nature
of the empirical distribution relative to a uniform distribution,
with $G = 0$ if all values of the variable have the same probability
of occurrence (``perfect equality'') while $G = 1$ corresponds to the
extreme situation with the variable always taking up a single value
(corresponding to a delta function probability distribution). 
Thus, if the
probability of occurrence of any sign (e.g., the letters
`A-Z' in the
case of the English alphabet) at the beginning (or end) of a word is about the
same, the corresponding Gini index will be close to zero. Otherwise,
it is a finite number ($\leq 1$) whose exact value depends on the extent of
inequality in usage of the different signs.
Measures
related to the Gini index have previously 
had limited use in the context of linguistic
sequences, e.g., to select
attributes for decision tree induction in classification for data
mining~\cite{Breiman1984}.

\begin{figure*}[tbp]
\begin{center}
\includegraphics[width=0.99\linewidth]{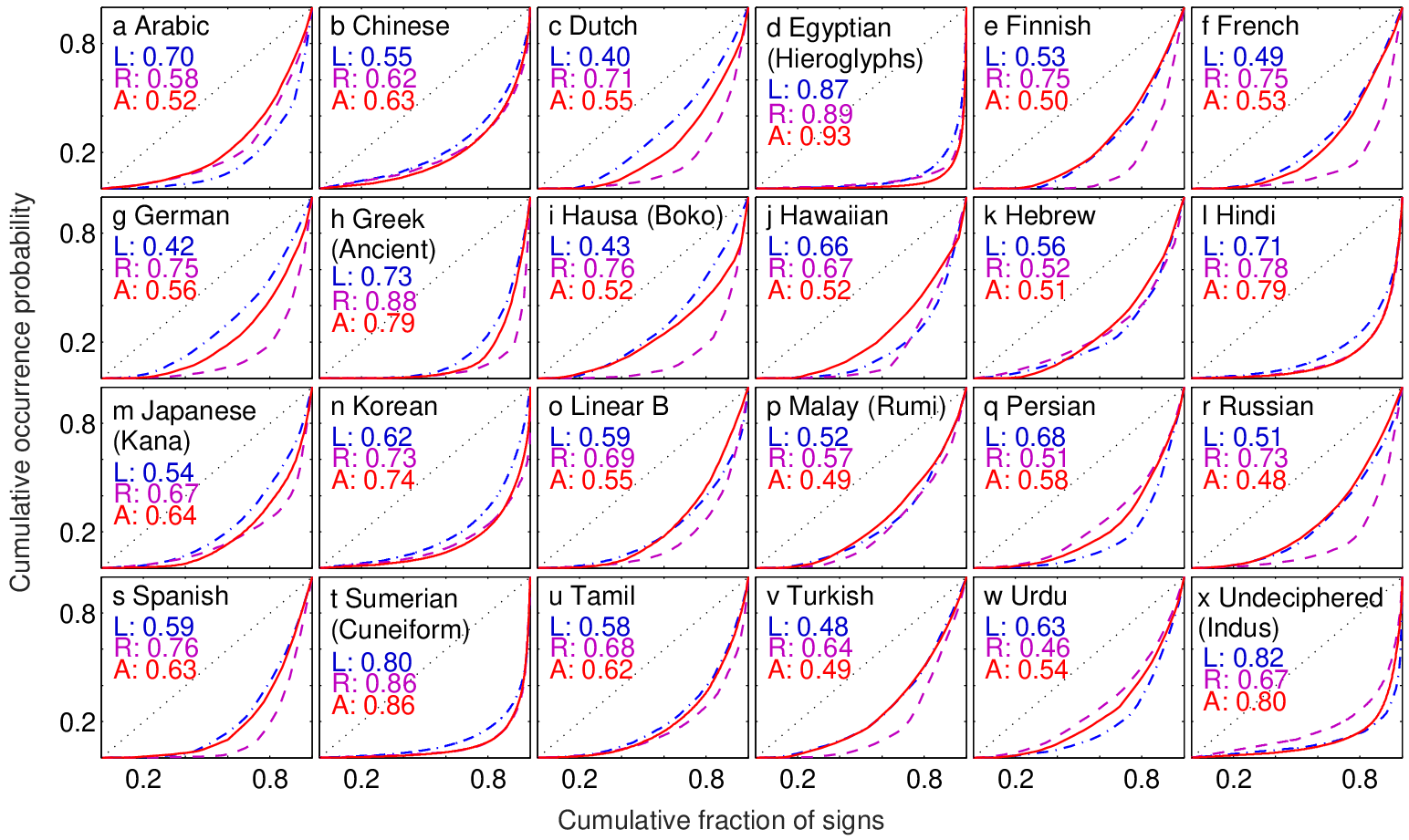}
\end{center}
\caption{
{\bf Unequal representation of signs (1-grams) occurring at different
positions in words in corpora written using different languages and
writing systems.} The Lorenz curves in the 24 panels (corresponding to
all the scripts analyzed here except English, which is shown in
Fig~\ref{fig1}) show the differences in the cumulative distribution function
of the occurrence probability of signs at left terminal position
(blue, dash-dot curve), right terminal position (purple, dashed curve) and at
any position (red, solid curve) of a word written in a particular
script.  The thin broken diagonal line corresponds to a perfectly
uniform distribution, deviation from which indicates the extent of
heterogeneity of sign occurrence distributions. This is measured in
terms of the Gini index,
the corresponding
values at the left terminal (L), right terminal (R) and any position
(A) for a script being indicated in each panel.
}
\label{fig2}
\end{figure*}
Using the Gini index on the distributions of letters (1-gram) 
that occur at the left and right terminal
positions of words in English, we can quantitatively express the
visible difference between 
patterns of unequal occurrence of signs at the two ends
seen in Figs~\ref{fig1}~(b-c).
Fig~\ref{fig1}~(d)
shows the Lorenz curve - a
graphical representation of the inequality of a distribution - for the
occurrence probability of the different 1-grams anywhere in a
sequence as well as the two terminal positions. 
For a set of $N$ symbols ($x_1$, $x_2$, \ldots $x_N$) 
that are indexed
according to their probability of occurrence in non-decreasing order
($P(x_i) \leq P(x_{i+1})$), the curve is obtained by joining using
linear segments the 
points ($X_i, Y_i$), $i = 1, \ldots N$, where
$X_i = i/N$ is the cumulative proportion of the population of symbols
and $Y_i = \Sigma^{i}_{j=1} P(x_j)/\Sigma^{N}_{q=1} P(x_q)$ is the
cumulative proportion of occurrence probabilities.
The diagonal line represents the case of complete
equality, so that higher inequality is manifested as greater deviation
between the empirical curve - showing the cumulative probability of
occurrence of signs arranged in a non-decreasing order of occurrence
frequency - and the diagonal.
The diagram clearly shows 
that for different signs the probabilities of
occurring at the right
terminal position, i.e., the end of a word, is more unequal
than their occurrence probability in the beginning (i.e., left
terminal position of a word), or indeed, 
anywhere in a sequence. This quantitatively establishes that there is
relatively more variation in the letters at the start of a
word - and conversely less so when ending it. It indicates an
inherent left-right asymmetry in the sign usage distribution of words
in English that is related to the different degrees of freedom
associated with choosing letters that begin and end a word.
This asymmetry is more pronounced for letters or 1-grams, as measured
by the normalized difference of Gini indices (defined
as $\Delta G = 2 (G_L - G_R)/(G_L + G_R)$ where $G_L$ and $G_R$ are the
Gini indices for the signs occurring in
the left and right terminal positions, respectively) 
for the two terminal
positions, $\Delta G = -0.50$ (with $95\%$ bootstrap confidence
intervals [$-0.48,-0.51$]; for details see Methods), than for
2-grams ($\Delta G = -0.25$, $95\%$ bootstrap confidence
intervals [$-0.247,-0.254$]) and becomes even less noticeable for
higher-order $n$-grams. We have therefore focused our
analysis on using
1-grams for the subsequent results reported here. 

To ensure that the left-right asymmetry does not arise simply as a
result of the use
of common prefixes (such as {\em de-} or {\em un-}) or suffixes (such as
{\em -ed} or {\em -ly}) in English, we have also analyzed the Ogden
list of Basic English words comprising two or more characters. This
is a set of commonly used English root words obtained after removing
all affixes or bound morphemes~\cite{Plag2002}.
For
this data-set, we obtain $\Delta G = -0.33$ (with $95\%$ bootstrap confidence
intervals [$-0.27, -0.47$]) clearly indicating that the relative lack
of variation in the right terminal position of a word in English is
not an artifact resulting from, say, a large number of words ending
with a limited set of suffixes.

\begin{figure*}[tbp]
\begin{center}
\includegraphics[width=0.99\linewidth]{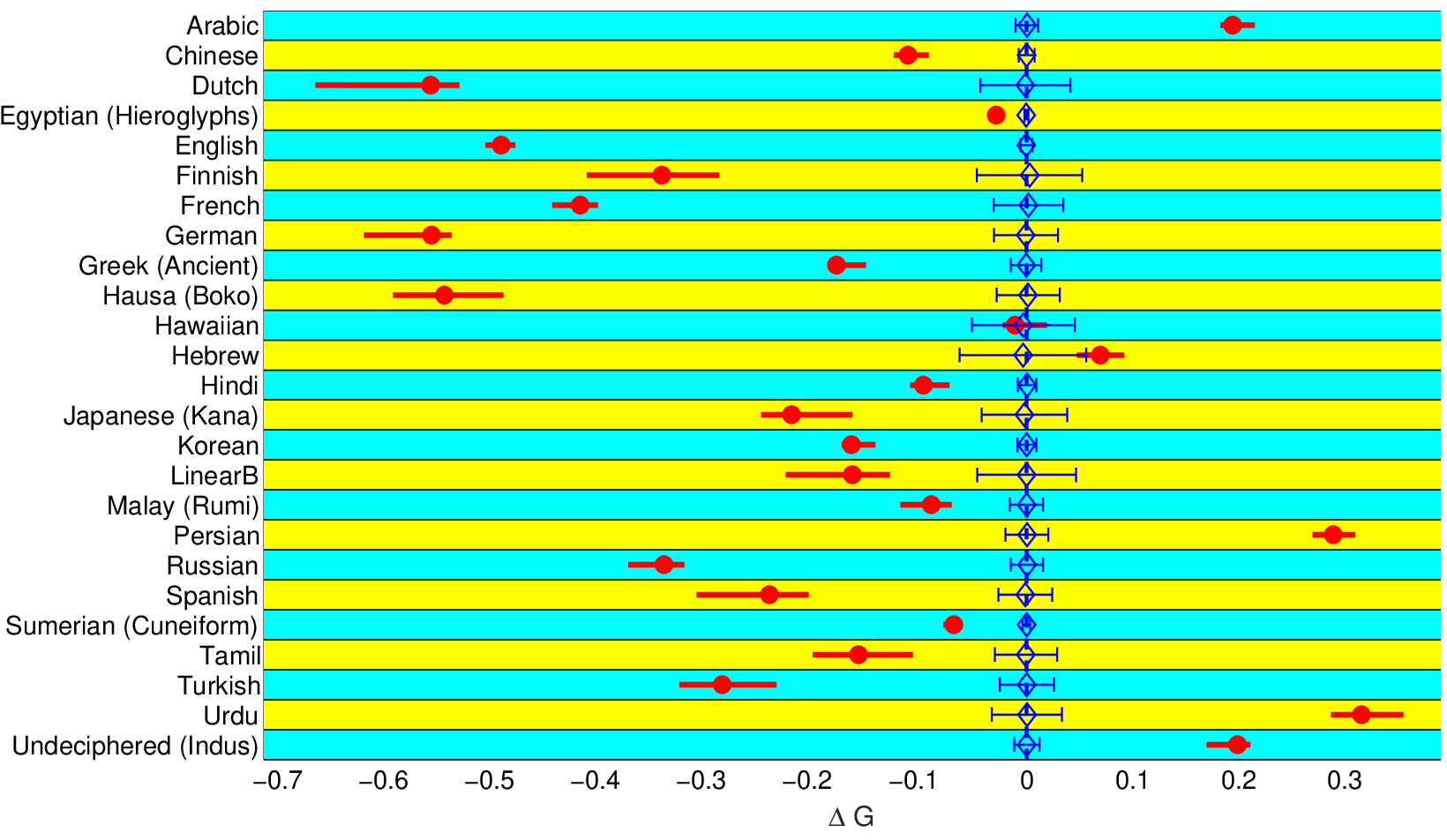}
\end{center}
\caption{
{\bf Asymmetry in the sign occurrence probability distributions at the
left and right terminal positions of words in different languages
correlate with the directions in which they are read.} The normalized
difference of the Gini indices $\Delta G = 2 (G_L - G_R)/(G_L + G_R)$
(filled circles), which measures the relative heterogeneity between
the occurrences of different signs in the terminal positions of words
of a language, are shown for a number of different written languages
(arranged in alphabetical order) that span a variety of possible
writing systems - from alphabetic (e.g., English) and syllabic (e.g.,
Japanese kana) to logographic (Chinese) [see text for details].  All
languages that are conventionally read from left to right (or rendered
in that format in the databases used here) show a negative value for
$\Delta G$, while those read right to left exhibit positive values.
The horizontal thick bars superposed on the circles represent the
$95\%$ bootstrap confidence interval for the estimated values of
$\Delta G$. To verify the significance of the empirical values, they
are compared with corresponding $\Delta G$ (diamonds) calculated using
an ensemble of 1000 randomized versions for each of the databases
(obtained through multiple realizations of random permutations
of the signs occurring in each word - see Materials and
Methods for details), the ranges of fluctuations being
indicated by error
bars. Along with the set of known languages, $\Delta G$ measured for a corpus of undeciphered inscriptions from the Indus Valley Civilization (2600-1900 BCE) is also shown (bottom row).
}
\label{fig3}
\end{figure*}

While this asymmetry in the usage distribution for signs that begin
and end words written in English is certainly striking, it would
be even more significant if the phenomenon turned out to be 
valid for linguistic sequences in general.
We have, therefore,
carried out a systematic investigation of the inequality in sign usage
distributions at the terminal positions of sequences that are chosen
from languages spanning a broad array of language families.
Fig~\ref{fig2} shows the Lorenz curves corresponding
to these different corpora, each indicating the differences in the
occurrence probabilities of signs at different positions for that
language.
The writing systems considered are also quite diverse, ranging from
alphabetic to logographic, whose corresponding signaries (i.e.,
the set of distinct characters used for writing in that system) can
vary in size from about two dozen to several thousands.
Fig~\ref{fig3} shows the results obtained for the different written
corpora we have analyzed, where the degree of asymmetry in sign occurrence 
at the left
and right ends of a sequence is measured by the
normalized difference $\Delta G$ between the respective Gini indices.
The most important feature of our results is the clear
distinction that can be made between languages that are conventionally 
written left to right, such as English, and those which are
written right to left, such as Arabic, according to the sign of $\Delta G$
obtained for the corresponding corpus. A negative value of $\Delta G$ 
implies that the signs occurring in the left terminal position have
a relatively more equitable distribution while the sign usage
distribution at the right terminal position is more unequal, and conversely for
positive $\Delta G$ comparatively few signs occur with high frequency at the
left end of a sequence than the right end.
Thus, our result implies that all languages and writing systems
considered here exhibit an asymmetry between the beginning and end of
a word in terms of the degree of inequality manifested in their
respective sign occurrence probability distributions, the probability in
choosing different signs being significantly more heterogeneous
at the end than in the beginning.

To ensure that the observed distinction between the sign usage
patterns at the two terminal positions of a sequence are significant,
we compared our results with those obtained from corpora of randomized
sequences, which by
design have the same distribution of sign occurrences at all positions.
For a rigorous comparison, we have used surrogate datasets that have
the same frequency distribution of different signs as the original
corpus (see Methods for details) so that any distinction between them
arises only from differences in the nature of the distributions of 
sign occurrence at the terminal positions.
As randomized sequences are expected not to have any left-right
asymmetry in sign usage patterns, the mean value of $\Delta G$ for
the surrogate data is expected to be zero. However, statistical 
fluctuations will result in the random corpora belonging to the
ensemble having small non-zero values of $\Delta G$ distributed about
$0$ and the standard deviation of the distribution (indicated by error bars in
Fig~\ref{fig3}) indicates whether a observed difference in Gini
indices can arise by chance even when there is no asymmetry.
As seen in Fig~\ref{fig3} almost all the corpora analyzed by us
exhibit asymmetries that are clearly distinct from what might be expected
if they were just
the result of noise. 

\begin{figure*}[tbp]
\begin{center}
\includegraphics[width=0.6\linewidth]{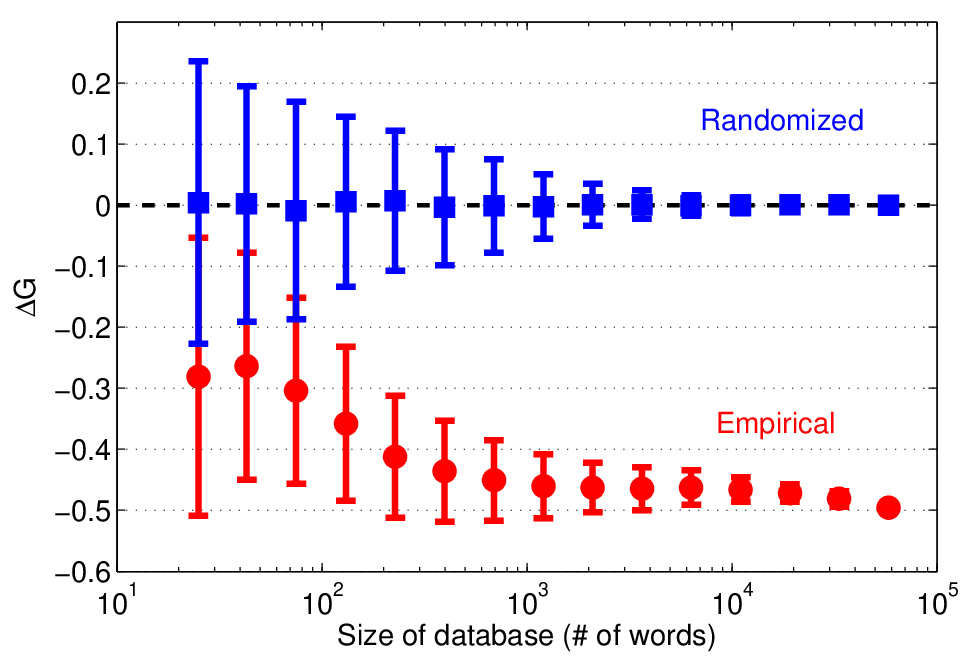}
\end{center}
\caption{
{\bf The observed asymmetry between heterogeneity of letter occurrence probability in left and right terminal positions is significant when the database is sufficiently large.} Gini index differential $\Delta G$ shown for the left and right terminal letter (1-gram) distributions calculated using a set of $N$ words, as a function of $N$. Empirical results are shown for random samples (without replacement) taken from the \textit{Mieliestronk} corpus comprising about 58000 unique words of the English language, each data point (circles) being the average over $10^3$ samples of size $N$. For each empirical sample, a corresponding randomized sample is created by randomly permuting the letters in each of the $N$ words, and a data point for the randomized set (squares) represents an average over randomizations of $10^3$ samples of size $N$. With increasing $N$ the empirical distribution becomes distinguishable from the randomized set (which, by definition, should not have any left-right asymmetry). The error bars indicate standard deviation over the different samples. 
}
\label{fig4}
\end{figure*}
As the asymmetry observed in the linguistic sequences should
not depend on the particular corpus from which they 
are chosen,
we obtained confidence intervals for the empirical $\Delta G$ values by
bootstrap resampling of the data (see Methods for details).
Fig~\ref{fig3} shows that in almost all cases this interval does not
have any overlap with the interval
obtained for randomized sequences -
indicating that our results are robust with respect to variations in
the corpus. The accuracy of the estimate, which is inversely related to the
length of the confidence interval, appears to become higher as the
database size, i.e., the total number of sequences being considered,
is increased. Indeed, Fig~\ref{fig4} shows that the database needs to 
be larger than a minimal size ($\sim 100$ words for English) in order
for the significance of the observed asymmetry to be established. 
Using increasingly larger databases, the difference between the
empirical and randomized corpora become more pronounced.

Apart from the Gini index, the inequality in sign occurrence distribution
at the terminal positions of a sequence can also be measured by other
means, e.g., using information entropy (also referred
to as Shannon entropy), a key concept in
information theory. 
It measures the amount of uncertainty in a process
by quantifying the non-uniformity in the probability
distribution of different events. Thus, if all signs
have equal probability of occurrence at a particular position in any
sequence,
this would correspond to the highest value of entropy (while the Gini
index would have been the lowest in this case). On the other hand, if
only a single sign occurs at this position in all sequences (i.e.,
the case of extreme inequality for which Gini index is highest), then the
entropy is zero.
Fig~\ref{figs2}
shows that using the normalized difference of entropy $\Delta S$,
estimated from the sign occurrence distributions at the left and
right ends of the different linguistic corpora, yields qualitatively
similar results to those obtained by using the Gini index. 
The sign of $\Delta S$ in almost all cases
is seen to be consistent with the direction of writing, with
left-to-right written languages having positive values of $\Delta S$
(with the lone exception of Egyptian Hieroglyphics, for which we have used
a database in which all sequences have been oriented so as to read
from left to right for all our analysis)
while those written right-to-left have $\Delta S <0$. This is in
broad agreement with our earlier conclusion that there is
relatively more equality in the probability of occurrence of different 
signs at the
beginning of a word than at its end - reflected in the higher
non-uniformity for sign usage distribution for the latter.
Thus, it suggests that the asymmetry we observe in
linguistic sequences may be robust with
respect to the specific measure of inequality being used.

Intriguingly, we find that the asymmetry can also appear in a corpus
of inscriptions that are so far undeciphered and whose relation to
language is therefore not yet established. As an illustration, we
have analyzed sign sequences appearing in the archaeological
artifacts (e.g., seals, sealings, pottery, copper tablets, etc.)
obtained from excavations carried out at sites of the 
Indus Valley Civilization (IVC) that existed during 2600-1900
BCE in present day Pakistan and
northwestern India~\cite{Possehl2002,Sinha2011b,Wells2015}.
While  there is some debate as to whether these inscriptions
constitute ``writing'' in the sense of encoding spoken
language~\cite{Lawler2004,Rao2009},
there is near unanimity among scholars that these were mostly
written from right to left as inferred from the archaeological
evidence (e.g., signs get more crowded at the
left end of some inscriptions or spill out of an otherwise linear
arrangement)~\cite{Hunter1934,Mahadevan1977,Parpola1994,Daniels1996}. We have used a database where the relatively few
sequences which are believed to have been written from left to right
have been reversed
so as to be oriented in the same direction as the majority, following 
standard procedure used for constructing concordances for Indus Valley
Civilization inscriptions.
We observe from Fig~\ref{fig3} that the $\Delta G$ for sign usage distribution
is positive, indicating that the choice of signs is less restricted in
the right terminal position than the left. This would suggest, based
on the connection previously seen between the sign of $\Delta G$ and
the direction of writing, that the IVC inscriptions are written from
right-to-left, which corroborates the consensus view as mentioned
above.

\section*{Discussion}
\label{sec3}
We have reported evidence here for a novel universal
feature in the empirical statistics of linguistic sequences.
Unlike the more well-known Zipf's law and Heap's law~\cite{Heaps1978}, 
which relate to
the frequency of word usage, we focus on a more elementary level,
viz., that of
the signs - corresponding to letters, syllabograms or logograms,
depending on the writing system - which constitute individual words.
The distribution of occurrence for the different signs at the left and
right terminal positions in a word are shown to
have a 
distinct heterogeneous character that are characterized by
measures of inequality such as the Gini index or information entropy.
We observe that, in general, the information content at the beginning
of a sequence tends to be higher than at the end, which is reflected
in the significant asymmetry in terms of the restriction of sign usage
at these the two positions. 
This is a pattern that is valid across different
languages and scripts, possibly revealing a feature inherent in 
the information processing and communicating capabilities of the human
cognitive apparatus.

The reason for the appearance of the directional asymmetry in sign
usage distributions for linguistic sequences is yet to be definitively
identified. However, it is not unreasonable to expect that this is
related to the phonotactic constraints inherent in different languages. The
initial sound of a word can be chosen with greater freedom from the
set of all available speech sounds (phonemes) of the language, compared to
all subsequent sounds that may depend - to a greater or lesser extent
- on the sound(s) preceding them. For example, very few of the
three-consonant clusters that can in principle occur in English are
actually allowed~\cite{McMahon2002}. Thus, one would expect a higher
degree of variability in the initial sound compared to the one at the
end of a word. As writing reflects the patterns of spoken language, to
greater or lesser extent depending on the system, one would expect
this difference between the beginning and end to be manifested in it. 
An indirect indication that
phonotactic considerations may be at least partially 
responsible for the asymmetry is
provided by the degree of the difference between the inequalities of
sign usage at the two ends of a word in different writing systems -
especially when normalized difference is used as a measure. We
observe that, broadly speaking, the magnitude of $\Delta S$ 
(as well as, $\Delta G$) is larger 
for scripts that have a higher proportion of phonetic
representation~\cite{DeFrancis1994,Robinson2009}.
Thus, alphabetic and syllabic systems which have a much greater
phonetic character than logographic or logo-syllabic systems tend to
typically show a more pronounced asymmetry~(Fig~\ref{figs2}). 
As even an apparently logographic system such as Chinese
has some
degree of phoneticism~\cite{DeFrancis1994}, it is not surprising that
systems having a high degree of logography also show a 
difference in the sign usage distribution between the beginning and end 
of words, although this
effect is much less marked than in other (more phonetic) scripts.
There are exceptions to this general trend - for example, Hebrew, which is
an alphabetic script, shows a low degree of asymmetry that is
difficult to distinguish
from effects due to stochastic fluctuations 
arising from sampling effects in a finite corpus. Hawaiian, which also
appears to have a very low $\Delta G$, however, shows a significant
asymmetry when information entropy is used to measure sign usage
inequality in place of the Gini index (see Supporting Information).
Other indications that a simple phonotactic explanation for the
observed asymmetry may not be adequate is shown by the fact that the 
relative position of some languages in terms 
of $\Delta G$ (or $\Delta S$) do not necessarily conform to common 
perceptions about the degree of phonetic representation in the
corresponding scripts used for writing them~\cite{Robinson2009}. 
For example, the Korean
han'g\u{u}l script is considered to have a higher proportion of 
phoneticism than French~\cite{Unger2004}; however, the latter exhibits
higher asymmetry in terms of both the measures of inequality used here
(see Fig~\ref{fig3} and Supporting Information).

The asymmetry reported here may be used to infer the direction of writing, which is one of the basic pre-requisites
for interpreting any linguistic sequence. A variety of possible
directions have been seen in different writing systems, both
historical and present~\cite{Coulmas1996}. 
The most common, left to right in horizontal
lines, is the direction in which all scripts descending from the Greek and
Br\={a}hm\={i} systems are written, including English, French, German,
Hindi and Tamil. Scripts that are written in the other direction,
i.e., right to left in horizontal lines, are also common and are used
in ancient and modern Semitic scripts including Arabic and Hebrew.
Another common orientation is from top to bottom in vertical columns,
which is the direction in which Chinese and scripts influenced by it
(such as Japanese) were traditionally written. 
Other, less common, directions
of writing are also known, including bottom to top (the Celtic Ogham
script) and {\em boustrophedon}, where the direction reverses in
successive lines (as in archaic Greek and Luwian hieroglyph
inscriptions). In cases where the inscriptions are undeciphered, such
as those of IVC, the direction usually has to be inferred by indirect means.
The asymmetry in sign usage patterns reported here - which shows that
the beginning of sequences can be distinguished from the end by the
nature of heterogeneity in the distributions of sign occurrence at
these positions - can provide a valuable tool for ascertaining the
direction of writing in such cases. 
Availability of a sufficiently large corpus would,
however, be necessary for a
reliable determination of the direction of writing in these
inscriptions.

\section*{Methods}
\label{sec4}

\bigskip
\noindent {\bf Data description.} 
We have analyzed data from written corpora of twenty four languages
(twenty two belonging to nine linguistic families, as well as, two
language isolates), along with a
corpus of undeciphered inscriptions from the Indus Valley Civilization
(ca. 2600-1900 BCE). The writing systems considered range
from alphabetic (that use only a few dozens of distinct letters) 
and syllabic to logo-syllabic and logographic
(involving thousands of signs).
The average corpus size is about ten thousand unique
words collected
from a variety of sources. Each word considered for our analysis
consisted of multiple graphemes, corresponding to letters, logograms,
hieroglyph signs or syllables depending on the writing system used.
Detailed description of each corpus is provided in the Supporting
Information.

\bigskip
\noindent {\bf Estimation of occurrence probability distribution.}
Probability distribution of sign occurrences in a corpus of
inscriptions are estimated from
frequency counts of the distinct signs appearing in the sequences
belonging to the database, i.e., 
$${\rm Prob(sign}~i~{\rm in~
position}~Q~{\rm)} = \frac{{\rm Number~of~sequences~where~sign}~i~{\rm
occurs~in~position}~Q}{{\rm Total~number~of~sequences}}.$$ 
For establishing the directional asymmetry
of sign usage, we focus specifically on the sign occurrence distributions 
at the left and right terminal positions of a sequence. The
inequality of sign usage at these positions, which is reflected in 
the non-uniform nature of the corresponding distributions, is
quantified by measuring the Gini coefficient or the information
entropy.

\bigskip
\noindent {\bf Measuring Gini coefficient.}
The Gini coefficient or index is a measure of how unequal are the
probabilities of all the different events that are possible.
A
value of zero for the coefficient corresponds to
situations where all events are
equally probable. Conversely, when only one event out of all
possible ones is observed in every instance, the Gini coefficient attains its 
maximum value of 1.
For a discrete probability distribution $P(x)$, where the $N$
possible values of the
discrete variable $x$ are indexed according to their
probability of occurrence in non-decreasing order ($P(x_i) \leq
P(x_{i+1})$, $i = 1, \ldots, N-1$), the Gini
coefficient~\cite{Brown1994} is measured as
\begin{equation}
G = 1 - (1/N) \Sigma_{i=1}^{N} [P_c (x_i)+P_c (x_{i-1})],
\label{ginieqn}
\end{equation}
where $P_c (x_i) = \Sigma_{j=1}^i P(x_j)$ is the cumulative probability of
$x$ with $P_c (x_0)=0$ and $P_c (x_N)=1$.
For a given set of inscriptions, we
estimate the cumulative probability
distributions $P_c^{(L)}$ and $P_c^{(R)}$ for the
signs $x_i$ occurring in the
left and right terminal positions, respectively, and use
Eq.~\ref{ginieqn} to compute the corresponding
estimated Gini indices $G_L$ and
$G_R$. 
The normalized difference between these two values, 
$$\Delta G = 2 (G_L - G_R)/(G_L + G_R),$$
provides a measure for the asymmetry in the
distribution of sign frequencies at the left and right terminal
positions of the sequences.
\bigskip

\noindent {\bf Estimating information entropy.}
Apart from Gini index, we have
used a measure based on information or Shannon entropy for quantifying the
nature of the inequality of sign usage distributions at left and right
terminal positions in a sequence. As entropy measures the
unpredictability of information generated from a source, it can be
used to characterize the underlying distribution of any process 
that produces a discrete sequence of symbols (chosen from a set of $N$
possible ones) and is defined as
\begin{equation}
S = - \Sigma_{i=1}^N P(x_i) \log_2
(P(x_i)),
\label{eqentropy}
\end{equation}
where $P(x_i)$ is the probability of occurrence of the $i$-th symbol and
the use of base 2 logarithm implies that the entropy can be expressed
in units of bits~\cite{Shannon1948}.
In particular, given any database of inscriptions, we obtain
estimates for the
probabilities $P^{(L)}(x_i)$ and
$P^{(R)}(x_i)$ for a particular sign
$x_i$ 
from the corresponding signary
to occur in the left terminal and right terminal positions. 
After estimating these probabilities for all $N$ signs 
that occur in
the corpus of inscriptions, the left and right terminal entropies
($S_L$ and $S_R$, respectively) are calculated by using Eq.~\ref{eqentropy}.
The normalized difference between the two entropy values, $\Delta S =
2 (S_L - S_R)/(S_L + S_R)$, provides a measure of the degree of
asymmetry in sign frequency at the two ends of a sequence.
\bigskip

\noindent{\bf Bootstrap confidence intervals.}
To quantify the degree of robustness in the estimates obtained using
the empirical databases, we have used a bootstrap method to obtain
confidence intervals for the measured values. For each corpus we have created
$10^3$ resampled datasets (i.e., bootstrap samples) by random sampling 
with replacement,
containing the same number of sequences as the original dataset.
The probability distributions for sign usage at the terminal positions
are then calculated for every bootstrap sample.
Finally, a confidence interval for the normalized difference between
the left and right terminal Gini indices, $\Delta G$ (or, of information
entropy, $\Delta S$) for the corpus is computed using the
Bias-Corrected and accelerated (BCa) method
which adjusts for bias in the bootstrap sample
distributions relative to the actual sampling distribution~\cite{Efron1987}.
The BCa confidence interval adjusts the percentiles
of the bootstrap distribution of the parameter according to the
calculation of a {\em bias correction} coefficient and an {\em
acceleration} coefficient. The former adjusts for any skewness present
in the bootstrap sampling distribution (it is zero if the distribution
is symmetric). The latter coefficient adjusts for nonconstant
variances (if any) in the resampled data.
\bigskip

\noindent{\bf Sequence randomization.}
The statistical significance of the measured asymmetry in sign usage 
is measured by comparing the results obtained from the empirical
database with an ensemble of randomized surrogate sequence corpus.
Each ensemble is generated by taking each sequence in turn that
belongs to a database and doing a random permutation of the signs.
This reordering ensures that the frequency distribution of the signs in each
sequence (and thus, also the corpus) is unchanged in the randomized
set, although all correlations (that contribute to 2- and higher order
$n$-gram distributions) are disrupted. We then perform the same
calculations as for the original empirical data for measuring the
degree of asymmetric sign usage in left and right terminal positions
of these randomized sequences - which, by design,  are expected not to
have any asymmetry. Significant difference in the results
of the two datasets ensures that the measured asymmetry is not arising
from stochastic fluctuations.


\section*{Acknowledgments}
We thank P. P. Divakaran, Deepak Dhar, Iravatham Mahadevan, Shakti N.
Menon, Adwait Mevada, Ganesh Ramachandran and Chandrasekhar Subramanian for helpful discussions and suggestions. We also thank Bryan K. Wells for allowing the use of the IVC sequence database compiled by him. This work was partially supported by the IMSc PRISM project funded by the Department of Atomic Energy, Government of India. \\


%
%
%

\pagebreak

\section*{Supplementary Information}
\setcounter{figure}{0}
\renewcommand\thefigure{S\arabic{figure}}

\vspace{0.7cm}
\noindent
{\large \bf Description of the corpora}\\

\noindent
{\bf Arabic:} We have used a database of 14867 unique words
(that are represented using two or more characters) of
Classical (or Quranic) Arabic, a Semitic language written using a
consonantal alphabet or `abjad'~\cite{Coulmas2003}. The words are
obtained from {\em Tanzil}, an international project started in 2007
to produce a standard Unicode text for the Qur'an
(\textit{http://tanzil.net/download/}, accessed: 25th March 2015).
The signary comprises 36 signs, viz., 28 consonantal signs and 8
consonants with
diacritical marks indicating vowels. The words range in length from 2
to 11 characters, the average length being 5.39.

\noindent
{\bf Chinese:} We have used a database of 
13104 unique words ({\em c\`{i}}) of the Chinese language (belonging to the 
Sino-Tibetan language family) that are written using two or more signs ({\em z\`{i}})~\cite{Chao1968}
which have been obtained from a
public-domain online
{\em Chinese-English dictionary}
CC-CEDICT
(\textit{http://www.mdbg
.net/chindict/chindict.php?pahe=cc-cedict},
accessed: 8th January 2011). The database contains annotations
identifying idiomatic expressions and loan words, and indicates proper
nouns by capitalization of the corresponding English translation.
From the total set of 115430 words and phrases available in the
dictionary we have removed all single character words, idiomatic
phrases, variants of the same word, hyphenated compound words, 
proper nouns and loan words.
The Chinese writing system has been variously described as either
logographic~\cite{Coulmas2003}, or, an imperfect phonographic
system with additional logographic
attributes~\cite{DeFrancis1984,DeFrancis1989,Sproat2000}.
The signary for our data comprises 3691 distinct graphemes
corresponding to logograms (hanzi). The sign sequences range in length
from 2 to 9 signs, the average length being 2.46. Traditionally,
Chinese is written top to bottom in vertical columns shifting from
right to left; however, in modern times it is more frequently being
written left to right in horizontal lines, and this is the convention
used in our database. The words in our database are written using traditional
Chinese characters.

\noindent
{\bf Dutch:} We have used a list of the 10000 most commonly used words
in Dutch, a member of the Germanic branch of the Indo-European
language family,
from which we have chosen the 
9146 unique non-hyphenated words comprising
two or more characters. The data has been collected
from the {\em Wortschatz} website maintained by the University of
Leipzig
(\textit{http://wortschatz.uni-leipzig.de/Papers/top
10000nl.txt},
accessed: 22nd May 2015). The signary
used has 
31
distinct alphabetic characters comprising 21 consonants, 5
vowels, 3 vowels with diacritical
marks (acute accents or diaeresis), the digraph `ij' that is
considered as a letter in the Dutch language
and
an extra letter from the German alphabet (the {\em Eszett}).
The words range in length from 2
to 
25 letters, the average length being 
7.6.

\noindent
{\bf Egyptian (Hieroglyphs):} Ancient Egyptian,
a member of the Afro-Asiatic (Hamito-Semitic)
language family, is written using a mixed system (also referred to as
a
logoconsonantal system~\cite{Daniels1996r}) with
several hundreds of {\em hieroglyph}
signs that can represent logograms,
phonograms and/or determinatives~\cite{Coulmas2003}.
We have used as data 39933 unique sequences comprising two or more hieroglyph
signs of the
{\em Middle Egyptian
Dictionary} compiled by Mark Vygus (updated April 2015,
\textit{http://www.pyramidtext
sonline.com/
MarkVygusDictionary.pdf},
accessed: 22nd May, 2015). The hieroglyphic signs are represented
using the
Gardiner sign list numbering system~\cite{Gardiner1957}, the signary
for the database used by us comprising 1859 distinct signs.
The sign sequences range in length from 2 to 17 hieroglyphs, the
average length
being 5.12.
The conventional direction of reading hieroglyphic sequences is
``toward
the face of human or animal pictograms, i.e., the signs are turned
towards the beginning of the inscription''~\cite{Daniels1996r}. In the
database used by us all sequences have been oriented so as to read
from left to right.

\noindent
{\bf English:} We have used the \textit{Mieliestronk} list of 58109
distinct words (comprising two or more letters) of the English
language - belonging to the Germanic branch of the Indo-European
language family - that has been compiled by merging several different
word-lists
(\textit{ http://www.mieliestronk.com/wordlist.html}, accessed: 4th
December 2011).
The words vary in length from 2 to 22
letters, the average being 8.34. The signary consists of the 26 lower
case letters of the English alphabet.
The list excludes spellings that are considered to be non-British.
If a word  is hyphenated, it is listed in unhyphenated
form by removing the punctuation mark. The list contains some
multiword phrases that are in common usage, rendered as a single word.
Several words are included in both their singular and plural forms.
Note that this is the word list for English that is included in the 
Supplementary Data Sets.

\begin{figure*}[!h]
\includegraphics[width=0.95\linewidth]{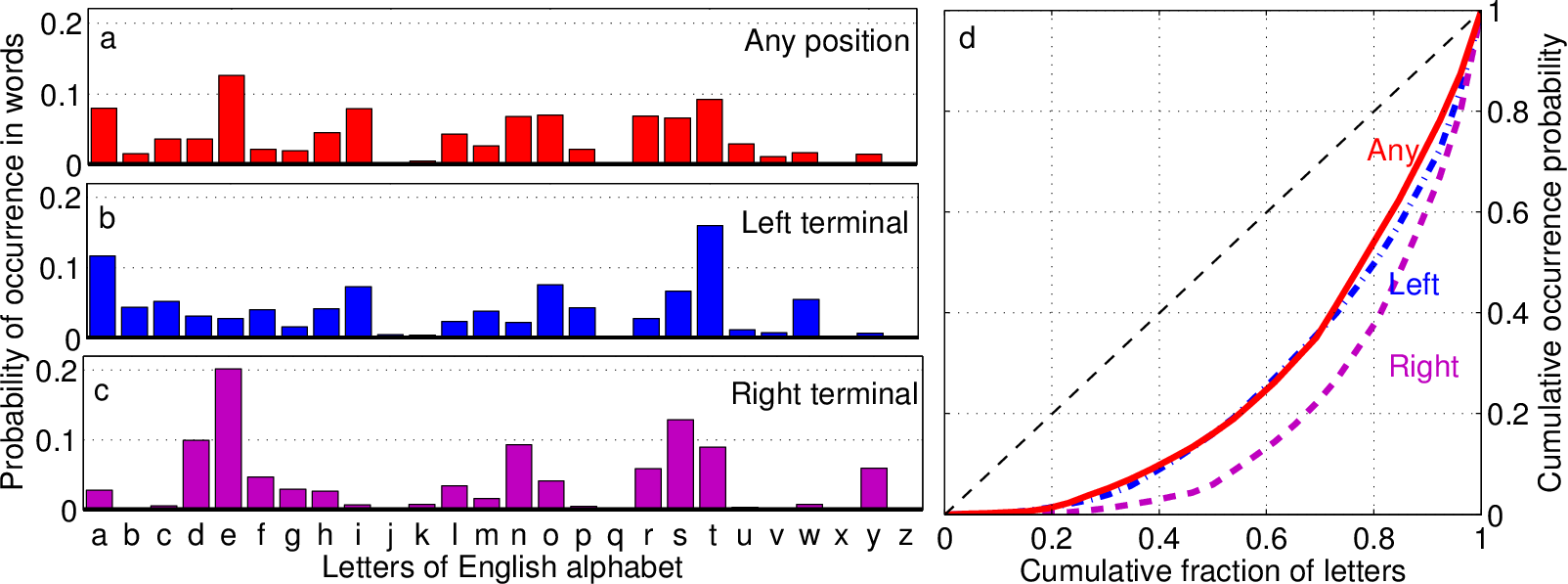}
\caption{
{\bf Unequal representation of letters (1-grams) occurring at
different
positions in words in written English is robust with respect to choice
of corpus.} The probability of occurrence
of the 26 letters of the English alphabet in the \textit{Google Books
Ngram} data
comprising about 97000 unique words of the English language that occur
with a frequency of more than 100,000 in the corpus (see
Methods for details),
at (a) any position, (b)
left terminal position (i.e., in the beginning) and (c) right terminal
position (i.e., at the end) of a word. While there are differences in
the occurrence probability of the individual letters with the
distribution shown in Fig~\ref{fig1} (see main text), as with the
{\em Mieliestronk} corpus there is higher
heterogeneity in (c) indicating that only a few
letters occur with high frequency at the right terminal position of a
word, compared to a relatively more egalitarian frequency of
occurrence
of letters in the left terminal position (b). This difference is
illustrated in the Lorenz curve (d) comparing the cumulative
distribution function for the occurrence probability of the different
letters in any (solid curve), left terminal (dash-dotted curve) and
right terminal position (dashed curve) of a word. The thin broken
diagonal line corresponds to a perfectly uniform distribution,
deviation from which indicates the extent of heterogeneity in 
the occurrence probabilities of different letters - measured as the ratio of the
area between the line of perfect equality and the observed Lorenz
curve, i.e., the Gini index.
}
\label{figs1}
\end{figure*}

For corroboration of our results obtained by analyzing the above
dataset, we have also used statistics of sequence
position-specific distributions of letter usage frequencies in a
list of 97565 distinct
words of the English language compiled from \textit{Google books
Ngram} data
(English Version 20120701, in
\textit{http://storage.googleapis.com/
books/ngrams/books/datasetsv2.html})
by Peter Norvig
and made freely available for public access
(\textit{http://norvig
.com/tsv/ngrams-all.tsv.zip}, accessed: 22nd
June 2015) \cite{Norvig2013}. Only those words are used which occur
with a frequency of more than
$100,000$ in the corpus of books scanned by Google.
Details of the procedure used for compiling the
N-gram frequency statistics are given in \textit{http://norvig.com
/mayzner.html} (accessed: 22nd June 2015).
Note that, there are some differences in the relative frequency of occurrence of the
different letters in the two databases. 
This is because, unlike 
in the preceding case where the occurrence probabilities are computed
from a database
comprising unique words, the Google 1-gram distributions are computed
from corpora of books containing multiple occurrences of the same word.
The frequencies of each letter is thus weighted by the occurrence frequencies in the corpus
of different words (whose distribution follow Zipf's law) containing that letter.
However, despite these differences in details, the
inequality of sign usage at the right terminal positions is visibly
higher than that in the left terminal positions (Fig~\ref{figs1}).

To ensure that our results are not an artifact of commonly used
affixes (e.g., prefixes like {\em de-} or {\em un-} and suffixes like
{\em -ed} or {\em -ly})
in English, we have also used the list of 850 words proposed by C. K.
Ogden~\cite{Ogden1930} as the core words of Basic English, 
a simplified subset of
regular English. This word list is widely used as the beginner's vocabulary
for the teaching of English as a second language. It comprises
only word roots, which for regular use are extended with different
affixes.
We have used the 848 words comprising two or more letters (i.e.,
omitting the single character words {\em a} and {\em i} from the Ogden
list), which range in length between 2 and 14, the average
length being 5.2.

In addition to these three data sets, we have tested several others of
varying sizes including a lemmatised list of frequently used words
from the British National Corpus~\cite{Kilgarriff1997}, lemma being
the canonical form of a word chosen to represent all the forms having
the same meaning. All of these data sets exhibit similar left-right
asymmetry as the above.

\noindent
{\bf Finnish:} We have used a list of the 10000 most commonly
used words (all of which use two or more letters)
in the Finnish language, belonging to the Finnic branch of the
Uralic language family.
The data, obtained from the \textit{Wikiverb} website, has been
collected from
newsgroup discussions, press and modern literature
(\textit{http://wiki.verbix.com/Documents/WordfrequencyFi}, accessed:
24th June 2015).
The signary used has 25 distinct signs - i.e., all vowels and
consonants of the modern Latin alphabet excepting ``q'',``x'' and
``w'',
along with two additional vowels ``\"{a}'' and ``\"{o}''.
The words vary in length from 2 to 25 letters, the average being 7.81.

\noindent
{\bf French:} We have used a list of the 10000 most commonly used
words in French, a Romance language belonging to the Indo-European
family, from which we
have chosen the 
9189 
unique words that have two or more characters. The data
has been collected from the \textit{Wortschatz} website
maintained by the University of Leipzig
(\textit{http://wortschatz.uni-leipzig.de/Papers/top10000fr.txt},
accessed: May 22nd 2015). The signary used has 30 distinct alphabetic
characters comprising 26 letters of the Latin alphabet along with 3
vowels with diacritical marks (acute accents or diaeresis)
and an apostrophe sign.
The words range in length from 2 to 19 letters, the average length
being 7.7.

\noindent
{\bf German:} We have used a list of the 9172 most commonly used
words in German, a member of the Germanic branch of the Indo-European
language family, from which we have chosen the 
9152 distinct words that have
two or more characters. The data
has been collected from the \textit{Wortschatz} website
maintained by the University of Leipzig
(\textit{http://wortschatz.uni-leipzig.de/
Papers/top10000de.txt},
accessed: May 22nd 2015). The signary used has 32 distinct alphabetic
characters comprising the 26 letters of the Latin alphabet
along with 4 vowels having diacritical marks (umlauts or acute
accents), a ligature (the {\em Eszett} or {\em scharfes S}) and
an apostrophe sign.
The words vary in length between 2 to 27 letters, the average being
8.1.

\noindent
{\bf Greek:} We have used a list of the 10000 most frequently
occurring words - grouped by lemma - in classical Greek literature
written in ancient Greek which belongs to the Indo-European language
family,
compiled by Kyle Johnson from the \textit{ Thesaurus Linguae Graecae}
corpus
(maintained by University of California, Irvine)
using the \textit{Classical Language} Toolkit
(\textit{http://cltk.org})
and made freely available to the
public
(\textit{http://kyle-p-johnson.com/assets/most-common-greek-words.txt},
accessed: 24th June 2015). From this dataset we have used the 
9868 distinct words that have two or more characters.
The signary has 124 distinct
characters as the words are represented in the traditional polytonic
orthography used for ancient Greek, involving 24 basic letters used in
conjunction with several varieties of diacritical marks (e.g.,
accents,
breathing marks, iota subscript and diaeresis).
The words range in length from 2 to 18 characters, the average being
6.9.

\noindent
{\bf Hausa (Boko):} Hausa, a Chadic language belonging to the
Afro-Asiatic family, is written using Boko, a Latin-based alphabet,
which was devised in the 19th century and became the official system
in the early part of the 20th century (in earlier periods, it was
written in  Ajami, an Arabic alphabet). We have used a list of 
7062 unique
words that have two or more characters obtained from a Hausa online
dictionary maintained by the University of Vienna
(\textit{http://www.univie.ac.at/
Hausa/KamusTDC/CD-ROMHausa/KamusTDC/ARBEIT2.txt},
accessed: 19th May, 2015).
The signary used has 30 distinct alphabetic characters comprising 23
letters from the Latin alphabet, four additional signs
representing glottalized consonants, two digraphs (`sh' and `ts')
and an apostrophe sign.
The words range in length from 2 to 22 characters, the average being
6.0.

\noindent
{\bf Hawaiian:} Hawaiian is Polynesian language belonging to the
Austronesian family that had no written form until the 19th century
when foreign missionaries devised an alphabetic system for recording
it based on the Latin script. The data used for our analysis has been
collected from the entries of
{\em A dictionary of the
Hawaiian language} (1922) compiled by Lorrin Andrews and revised by
Henry
H Parker (Board of Commissioners of Public Archives of the Territory
of Hawaii, Honolulu) and freely available online
(\textit{http://ulukau.org/elib/cgi-bin/library?c=parker\&l=en},
accessed on 28th May 2015).
After removing all non-native words that contain
characters that do not belong to the Hawaiian alphabet, we have
compiled a data-base of
14009 unique words containing two or more characters.
The signary comprises 28 distinct characters, with 12 basic letters -
representing 5 vowels and 7 consonants -
of the Hawaiian alphabet
along with vowels used in conjunction with diacritical marks (breve
and macron)
indicating short or long pronunciation, and a sign to indicate glottal
stop
(the {\em `okina}).
The words range in length from 2 to 26 characters, the average being
9.0.

\noindent
{\bf Hebrew:} We have used a list of the 10000 most commonly used
words (compiled from online written texts) in modern Hebrew, a Semitic
language written using a consonantal alphabet or `abjad', from which
we have chosen the 9993 distinct words that are
represented using two or more characters.
The data has been collected from a list maintained by {\em Teach Me
Hebrew}, an online Hebrew language learning site
(\textit{http://www.teachmehebrew.com/hebrew-frequency-list.html},
accessed: 26th December 2013).
The signary comprises 31 signs, viz., 27
consonantal signs (comprising 22 letters of which five use different
forms -
called {\em sofit} - when used at the end of a word) and 4 signs used
in
conjunction with {\em niqqud} diacritical
marks. The words range in length from 2 to 13
characters, the average length being 5.1.

\noindent
{\bf Hindi:} Hindi is an Indo-Aryan language, a branch of the
Indo-European family, which is written in the Devanagari script that
is sometimes classified as an alphasyllabary~\cite{Daniels1996r} or
`abugida'~\cite{Daniels1990}.
Like the other writing systems that are descended from the
Br\={a}hm\={i} script of ancient India, Devanagari uses
as its main functional unit the {\em aksara}, which may consist of
only a vowel but more frequently represents
a syllable
consisting of a consonant and an inherent vowel along with diacritical
marks
that may indicate use of other vowels~\cite{Coulmas2003}.
We have used a database of 6441 distinct words written using two or more
characters, collected from an online
dictionary ({\em Shabdanjali}) of Hindi developed by the
Language Technology Research Center at Indian Institute of Information
Technology, Hyderabad
(\textit{http://ltrc.iiit.ac.in/showfile.php?filename=downloads/
shabdanjali-stardict/index.html},
accessed: 20th May 2015).
The signary comprises 571 distinct signs, comprising 11 vowels, 33
consonants,
their conjunctions with each other
and with consonantal sound modifiers (the {\em
anusvara}, {\em chandrabindu}, {\em visarga} and {\em halant}).
The words range in length from 2 to 10 signs, the
average length being 3.5.

\noindent
{\bf Japanese (Kana):} Japanese, which belongs to the Japonic
language family, is written using a combination of the logographic
{\em Kanji} system (adopted from Chinese characters) and the syllabic
{\em kana} system. The latter, in turn, consists of a pair of distinct
syllabaries: {\em hiragana}, used for writing native Japanese words
and
{\em katakana}, which is used for foreign words. For our study, we
have focused only on the syllabic writing system for Japanese.
We have used a list of 
1162 distinct words written using two or more
signs from the kana syllabary, which is obtained from a list
of common Japanese words collected from textbooks used by foreign
learners of the language and
maintained by {\em Japanese Words}, an online site for learning
the Japanese language
(\textit{http://www.japanesewords.net/36/over-1000-japanese-words-list/},
accessed: 29th May 2015).
The signary has 103 distinct characters comprising 46 basic signs of
Hiragana
and 21 basic signs of Katakana, 22 Hiragana and 9 Katakana signs
used in conjunction with diacritical marks (the {\em dakuten} and {\em
handakuten}), smaller forms of 4 hiragana characters
(viz., of {\em ya}, {\em yu} and {\em yu} which
indicate the y\={o}on feature, and the {\em sokuon} used to mark a
geminate
consonant) and a special symbol ({\em ch\={o}onpu}, the long vowel
mark).
The words range in length from 2 to 13 characters, the average being
3.8.

\noindent
{\bf Korean:} Korean is a language isolate with no established
connection to any of the major language families of the world and is
written using Han'g\u{u}l, a purely phonetic script, although in
earlier times a system based on Chinese characters (Hanja) was used.
Each character corresponds to a syllable, the syllabic block being
composed of two to six letters (including at least one consonant and
one vowel) from the basic alphabet comprising 10 vowels, 14 consonants
and 27 digraphs. The number of possible distinct syllabic blocks or
characters exceeds 11,000 although a far smaller number is in actual
use~\cite{Coulmas2003}. We have used a list of 5888 commonly used
words compiled by the National Institute of Korean Language in 2004
(publicly accessible from
\textit{https://en.wiktionary.org/wiki/Wiktionary:Frequency\_lists/
Korean\_5800}, accessed: 24th June 2015)
from which we have chosen 
5141 distinct words that are represented using
multiple
syllabic blocks.
The signary comprises 
966 distinct characters (each corresponding to
a syllabic block). The words range in length from 2 to 6 characters,
the average length being 2.7.

\noindent
{\bf Linear B:} Linear B is a syllabic script, with most of its signs
representing consonant-vowel combinations,
that was used for writing archaic Greek between 1500 and 1200 BCE.
We have used as data 
1924 distinct sequences comprising two or more characters
from the {\em Linear B Lexicon} compiled by Chris Tselentis
(\textit{https://www.scribd.com/doc/56265843/
Linear-B-Lexicon},
accessed: 15th May 2015).
The signary comprises 87 distinct signs representing syllables.
The words range in length from 2 to 8 signs, the average being 3.8.

\noindent
{\bf Malay (Rumi):} We have used a list of the 10000 most commonly
used words in Malay, a member of the Austronesian language family,
from which we have chosen the 9970 unique words that
have two or more characters. All words are written in {\em Rumi}
or Latin script, which is the most commonly used form for writing
Malay at present, although a modified Arabic script ({\em Jawi}) also
exists.
The data has been collected from the list of high frequency words
that are publicly available at {\em Invoke IT Blog}
(\textit{https://invokeit.wordpress.com/frequency-word-lists/},
accessed: 4th January, 2014).
The signary comprises the 26 letters of the Latin alphabet.
The words range in length from 2 to 17 letters, the average being 6.8.

\noindent
{\bf Persian:} We have used a list of 10000 most commonly used words
(each represented using two or more characters) in Persian, a member of
the Indo-Iranian branch of the Indo-European language family, which is
written using a modified form of the consonantal Arabic alphabet or
`abjad'. The
words are obtained from a list of high-frequency words compiled using
the Tehran University for Persian Language corpus and available at
{\em Invoke IT Blog}
%
(\textit{https://invokeit
.wordpress.com/frequency-word-lists/},
accessed: 4th January 2014).
The signary comprises 40 signs, viz., 32 consonantal signs, a long
vowel indicator (`alef madde'), a ligature
(`l\={a}m alef'), a diacritic (`hamze'), 3 consonants with the
`hamze' diacritical mark and different forms
for the consonants `k\^{a}f' and `ye' when they occur in final
position.
The words range in length from 2 to 13 letters, the average being 5.2.

\noindent
{\bf Russian:} We have used a list of 
9011 distinct words that use two
or more characters in Russian, a member of the Slavic branch of the
Indo-European language family and which is written using a Cyrillic
alphabet. The data has been collected from
{\em Russian Learners' Dictionary: 10,000 words in frequency order}
compiled by Nicholas J Brown (Routledge, London, 1996), after removing
all words that use characters not in the standard Russian alphabet.
The signary comprises the 33 letters of the modern Russian alphabet.
The words range in length from 2 to 21 letters, the average being 8.0.

\noindent
{\bf Spanish:} We have used a list of 
4902 distinct high-frequency words (that
use two or more characters) in
Spanish, a Romance language belonging to the Indo-European family.
The data has been collected from {\em A Frequency Dictionary of
Spanish} compiled by Mark Davies (Routledge, London, 2006).
The signary used has 35 distinct alphabetic characters comprising 26
letters of the basic Latin alphabet along with an additional character
\~{n} and two digraphs (`ch' and `ll'), as well as, vowels with
diacritical
marks (acute accents or diaeresis).
The words range in length from 2 to 19 letters, the average being 7.4.

\noindent
{\bf Tamil:} Tamil is a Dravidian language is written in a script
(sometimes classified as an `abugida')
derived from Br\={a}hm\={i} script and thus shares
a common origin with the Devanagari script used for writing Hindi (see
above) although it differs significantly both in appearance and
structure~\cite{Coulmas2003}. It has 31 basic signs consisting of 12
vowels, 18 consonants and a special character, with combinations of
the different vowels and consonants yielding a possible
216 compound letters.
Additional characters from the Grantha
script and diacritical marks
are sometimes used to represent sounds not native to Tamil, e.g., in
words borrowed from other languages.
As with other Indian scripts, Tamil uses the {\em aksara}
as its basic unit - however, unlike then it has eliminated most
conjuncts, consonant clusters being placed in a linear string.
The data used for our analysis has been collected from
texts (e.g., Paripaadal, Thiruppavai, Kamba
Ramayanam, Sundara Kandam, Akananooru songs, etc.) available in
{\em Chennai Library}, an online repository of Tamil literature
(\textit{http://www.chennailibrary.com}, accessed: 27th December
2012).
From this a data-base of 1991 unique words containing two or more
characters was compiled. The signary comprises 187 distinct signs
corresponding to different basic and compound letters and the special
character.
The words range in length from 2 to 9 letters, the average being 3.8.

\noindent
{\bf Turkish:} We have used a list of 
9909 distinct high-frequency words (that
use
two or more characters) in Turkish, a member of the Turkic language
family. The data has been collected from a {\em Wiktionary} word
frequency list
(\textit{https://en.wiktionary.org/wiki/
Wiktionary:Frequency\_lists/
Turkish\_WordList\_10K}, accessed: 14th July 2015).
The signary used has 32 letters, comprising 29 letters of the Turkish
alphabet and 3 vowels used in conjunction with circumflex
accents.
The words range in length from 2 to 17 letters, the average being 6.9.

\noindent
{\bf Sumerian (Cuneiform):} Sumerian, a language isolate spoken in
ancient Mesopotamia
during the 3rd millennium BCE, was written using a logosyllabic system
with several hundred signs of a cuneiform script representing
logograms, phonograms and/or determinatives. We have used as data
19221 unique words (comprising two or more cuneiform signs)
collected from texts available in the {\em Electronic Text Corpus of
Sumerian
Literature} (ETCSL, \textit {http://etcsl.orinst.ox.ac.uk}, accessed:
20th October 2010).
The signary comprises 1364 distinct transliteration values of
Cuneiform signs that represent syllables.
The sequences range in length from 2 to 11 signs, with the
average being 3.6.

\noindent
{\bf Urdu:} We have used a database of
4998 unique words that are represented using two or more characters
in Urdu, an Indo-Aryan language belonging to the Indo-European family,
that is written using an extended Persian alphabet. The words are
obtained from a list of frequently used words maintained by the
Center for Language Engineering at Lahore
(\textit{http://www.cle.org.pk
/software/ling\_resources/UrduHighFreqWords.htm},
accessed: 1st January 2014).
The signary comprises 46 signs, viz., 38 consonantal signs, 3 long
vowels (`alef madde', `l\={a}m alef madde' and `ya'), 2
semi-consonants (`hamzah' used in conjunction with `wao' or `ya'), a
nasalized consonant (`noon ghunna'), a ligature (`l\={a}m alef') and
an additional sign (`ta' marbuta') used for writing certain
loan-words.
The words range in length from 2 to 11 letters, the average being 4.6.

\noindent
{\bf Undeciphered (Indus):} As an example of an undeciphered corpus on
which to apply our analysis, we have used the set of
inscriptions obtained from archaeological excavations at various sites
of the Indus Valley civilization (ca. 2600-1900 BCE). The data used
for our analysis is collected from the ICIT Database of Indus Writing
compiled by Bryan K. Wells~\cite{Sinha2011br,Wells2015r} and
maintained by Andreas Fuls
(\textit{http://caddy.igg.tu-berlin.de/indus/welcome.htm}, accessed:
20th October 2017)
from which we
have removed all incomplete and multiple-line inscriptions thereby
obtaining
1837 unique sequences that contain two or more signs.
The Indus signs are represented using the Wells sign list
numbering system~\cite{Sinha2011br,Wells2015r}, the signary for the
database used by us comprising 
568 distinct signs.
The sequences range in
length from 2 to 13 signs, the average being 4.6.
The direction of the sign sequences vary, the majority being written
right to
left, although examples of left to right 
also exist, 
as inferred from external evidence such as signs becoming relatively
cramped towards the end of a sequence inscribed on an archaeological 
artifact (e.g., seals, tablets or potsherds)~\cite{Mahadevan1977r,Parpola1994r}.
In the database used by us~\cite{Wells2015r}, the relatively few
sequences which are believed to have been written from left to right
have been reversed so as to be oriented in the same direction as the
majority (i.e., which are believed to be written from right to
left). This follows the standard procedure used also for constructing
earlier concordances for Indus Valley Civilization
inscriptions~\cite{Mahadevan1977r,Parpola1994r}.
Note that if only the 1779 sequences which are believed to be written from
right to left are considered for our analysis, 
we obtain $\Delta G = 0.20$ and $\Delta S = -0.41$ which are identical
to the values reported in the main text
for the database in which all 1837
sequences are oriented right-to-left.

\vspace{1cm}
\noindent
{\large \bf Robustness of results}\\

\noindent
An important consideration when quantifying the inequality of sign
usage is the size of the signary, i.e., the number
of visually distinct signs (sometimes referred to as
`graphs'~\cite{Coulmas1996r}) that can be identified in each corpus.
In several scripts, complex characters that are recognizably
the compound of two or more basic characters are quite commonly used -
as in the system of conjunct consonant signs or ligatures in the
Devanagari script used for writing Hindi and other South Asian
languages - and in principle, one could either consider these as
separate signs or decompose them into the constituent signs, which
will
result in very different signary sizes. Also, Semitic scripts such as
Arabic and Hebrew that are essentially consonantal alphabets often use
diacritical marks for indicating vowel usage. In such cases, the
same consonant is used in conjunction with different diacritics when
the vowel following it is different. The signary size would depend
upon
whether these are considered to be distinct signs or not. In several
other scripts, special marks can be used together with the vowels
and consonants, e.g., the use of apostrophe to indicate the
omission of one or more sounds in European languages such as French or
German, and the use of a glottal stop marker ({\em 'okina}) in
Hawaiian. Once again, whether these signs are treated as distinct
elements or part of the associated letter will
affect the signary size. We observe that although the numerical values
of the Gini indices (and information entropy) can be affected by
changes in the signary
size, the asymmetry in terminal
sign usage reported here is robust with respect to these choices about
different
conventions for identifying distinct signs constituting the signary
for a given corpus. 
\\
We have also examined our results for their
dependence on corpus size, i.e., the number of words included in each
database. As shown in Fig~\ref{fig4} in the main text (for English) and
Fig~\ref{figs3} in Supplementary Information (for Persian), the scores for
each language indicative of their asymmetry converge to a value which
is relatively independent of the corpus size as the number of words
included in the corresponding database is increased. We have
explicitly checked
that the scores for the different languages attain asymptotic values
(for large enough database sizes)
which are distinct from each other. Thus the variation of the scores
for the different languages cannot be attributed to corpus size alone.\\

\vspace{1cm}
\noindent
{\large \bf Possible non-phonotactic mechanisms for emergence of
asymmetry}\\

\noindent
It is possible that the asymmetry we have reported here could arise in
certain
situations for reasons other than phonotactic constraints. For
instance, in the undeciphered IVC inscriptions, the most frequently
occurring sign - viz., the U-shaped ``jar''
symbol~\cite{Possehl2002r} - that appears very frequently at the end of
a sequence,
dominates the probability distribution of signs that
can occur at the left terminal position (accounting for about a third
of all the distinct sequences comprising the corpus).
By contrast, the most frequently occurring sign at the right terminal
position is seen to begin only about $6~\%$ of the sequences.
These distinctive sign usage patterns at the two terminal
positions of IVC sequences gives rise to the heterogeneity in the
corresponding occurrence probability distributions. In the absence of
a decipherment, it is purely speculative whether the inequality arises
for phonotactic reasons (as in the linguistic sequences considered
here) or a fortuitous stylistic convention.\\

\vspace{1cm}
\noindent
{\large \bf Appearance of sign usage asymmetry in writing systems that
are not alphabetic or syllabic}\\

\noindent
One of the writing systems included in our study, that of Chinese language, is traditionally
considered to be logographic~\cite{Coulmas2003}. It is thus worth considering 
whether a non-phonotactic mechanism could be responsible for the appearance
of sign usage asymmetry in such a system. It has been reported that, while
single signs or characters ({\em z\`{i}}) can indeed represent words, the
majority of Chinese words ({\em c\`{i}}) 
are written using multiple signs, some studies suggesting as much as 70\%
of commonly used words involve two or more signs~\cite{Yip2000}.
While some of these words are polysyllabic, several others are compound words
created by affixation, i.e., combining a word with a prefix, infix or suffix.
As usage of suffix in Chinese is much more common than other types of affix 
usage~\cite{Li1981},
this may possibly be a mechanism by which sign usage asymmetry can arise
(e.g., the use of a limited set of suffixes at the end of a multi-character word
can reduce the diversity of sign usage at that position). However, we have
carried out analysis of the database of Chinese words after eliminating
words with affixes and observed that the sign usage asymmetry still persists.
A plausible hypothesis one may consider is that the remaining words in the database 
possess some degree of phonetic character. Thus, some variant of the phonotactic 
argument for the observed asymmetry suggested in the main text may arise even
for this system. We note that several authors have indeed argued that, instead of 
treating it as a purely logographic system~\cite{Coulmas2003}, Chinese writing is better thought 
of as an ``imperfect phonographic system with additional logographic attributes''~\cite{DeFrancis1984}. Indeed the importance of phonetic information in the development of the Chinese writing system is considered to be unequivocal~\cite{Sproat2000}. 
\\

We also note that other writing systems that use logograms, such as the Egyptian
hieroglyphic system, are nevertheless phonological in character to a considerable 
degree. Indeed, the recognition that Egyptian hieroglyphic writing system is significantly phonetic
was crucial to its decipherment by Jean-Francois Champollion~\cite{Bard2015}. 
Egyptian represents an apparently unique polyconsonantal system~\cite{Sproat2000},
with extensive use of determinatives that appear at the end of words (following the 
phonetic signs) supplying specified meanings.
It is intriguing to consider whether a variant of the phonological
argument that has been suggested to underlie the sign usage asymmetry in alphabetic
and syllabic systems could also apply to these apparently logographic and 
logoconsonantal systems.     

\vspace{1cm}
\noindent
{\large \bf Entropy difference in Egyptian hieroglyphs}\\

\begin{figure*}[ht]
\includegraphics[width=0.95\linewidth]{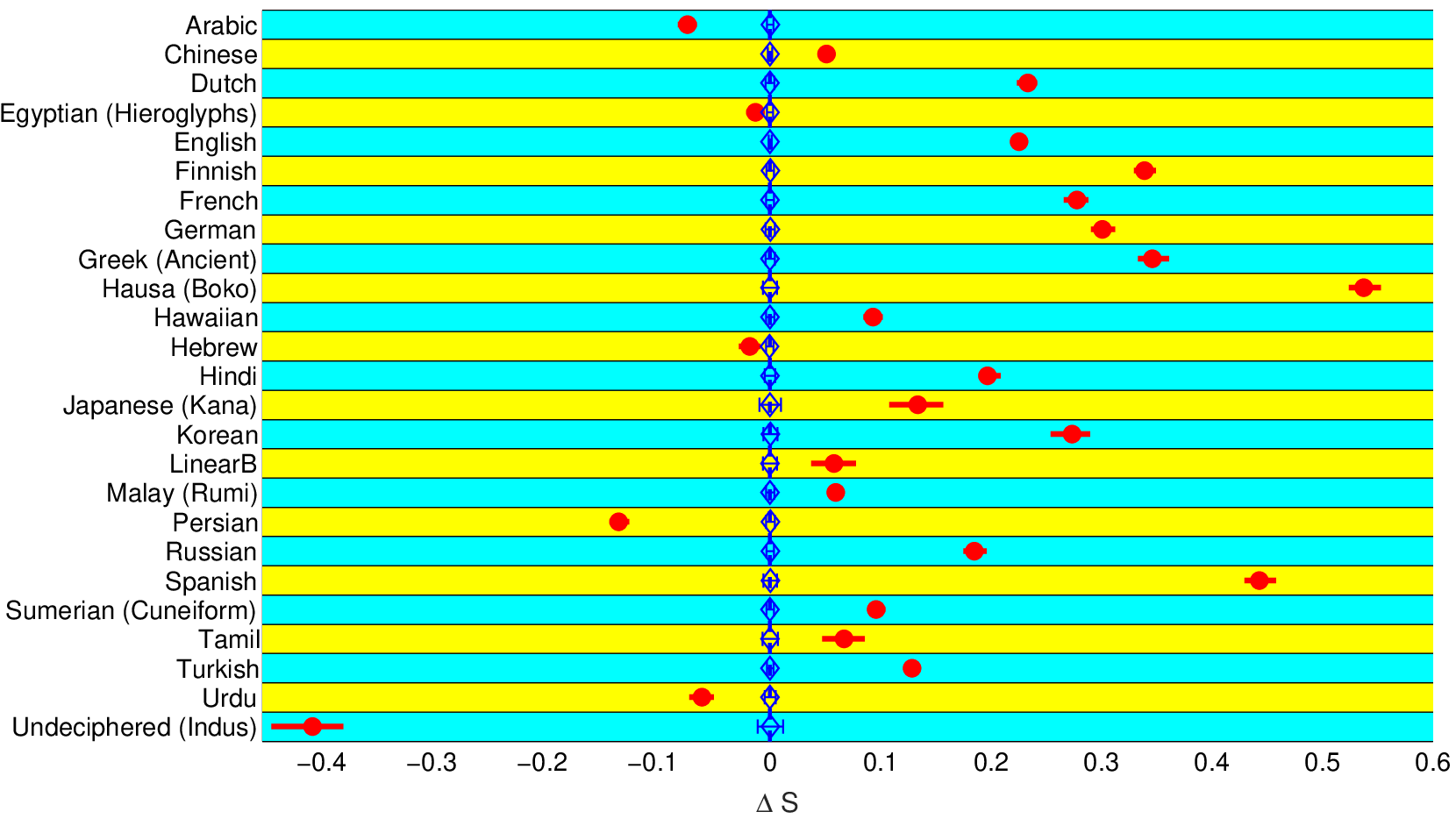}
\caption{
{\bf Asymmetry in the sign occurrence probability distributions at the
left and right terminal positions of words
in different languages appear to be relatively robust with respect to the quantitative
measure of inequality used.}
The normalized difference of the Shannon or information entropies
$\Delta S = 2 (S_L - S_R)/(S_L + S_R)$
(filled circles),
which measures the relative heterogeneity between the
occurrences of different
signs in the terminal positions of words of a language, are shown for
a number of different written languages (arranged in alphabetical
order) that span a variety of possible writing systems - from
alphabetic (e.g., English) and syllabic (e.g., Japanese kana) to
logographic (Chinese) [see text for details].
Almost all languages that are conventionally read from
left to right (or rendered in that format in the databases used here)
show a positive value for $\Delta S$ (with the exception of Egyptian hieroglyphs), while those read right to left
exhibit negative values. The horizontal thick bars superposed on the
circles represent the bootstrap confidence interval for the estimated
values of $\Delta S$.
To verify the significance of the empirical values, they are compared
with corresponding $\Delta S$ (diamonds)
calculated using an ensemble of randomized
versions for each of the databases (obtained through multiple
realizations of random permutations of the signs occurring in each
word). Data points are averages over 1000 random realizations, the
ranges of fluctuations being indicated by error bars.
Along with the set of known languages, $\Delta S$ measured for a
corpus of undeciphered inscriptions from the Indus Valley
Civilization (2600-1900 BCE) is also shown (bottom row).
}
\label{figs2}
\end{figure*}

\noindent
As can be seen from Fig~\ref{figs2}, for almost all sequence databases that
are read from left to right (or rendered in that format) we see a
positive value of $\Delta S$ except for Egyptian hieroglyphs. Instead
of having a higher entropy for the left terminal sign compared to
right terminal sign (as should have been the case given that all
sequences have been oriented to be read from left to right in the
database), it exhibits a marginally higher entropy for right terminal
sign in comparison to left terminal sign. Note that the numerical
value of $\Delta S$ for this data-set lies almost within the error
bars of the corresponding randomized data-set, so that the direction
cannot be conclusively determined given the data. The discrepancy may
also be connected to a distinctive feature of Egyptian hieroglyphic
inscriptions, viz., there are many more rare or low-frequency signs
that can appear in the right terminal position but never in the left
terminal position. Thus, there are 1502 distinct signs that appear in
the right terminal position of a sequence while only 798 distinct
signs appear in the left terminal position. If we compute the entropy
of the characters occurring in the left and right terminal positions
by confining our attention to the $r$ most frequently occurring signs
for either position (so that the number of terms used for computing
the two entropy values are comparable) we note that the left terminal
sign entropy is consistently higher than the right terminal sign
entropy for all $r$ (see Fig~\ref{figs4}). This is consistent with our hypothesis according to which the left terminal position should have exhibited higher entropy. 

\vspace{1cm}
\noindent
{\large \bf Inequality of sign usage distributions at different
positions in a sequence}\\

\noindent
In the main text we have only considered the difference in the occurrence probability of signs
at the two extreme positions of a sequence, viz., at its beginning and at its end. However, one can also
consider the sign usage distribution at other positions in a sequence, and in particular,
ask whether there is a systematic variation in the inequality measures of sign usage frequencies
with the relative position of their occurrence in a sequence.
To study this, instead of considering the entire data-set of sequences in a given language (e.g., English words) together, we split it into sub-sets, each having sequences of the same length. This is necessary as otherwise, it is difficult to compare the properties of (say) the 4th character in a sequence when the sequence itself is a 4-letter word (in which case, the character is also the right terminal symbol) and the 4th character in a 7-letter world (in which case it is in the middle of the sequence). 

We show in Figs~\ref{figs5} and \ref{figs6} the variation of the Gini index $G$ and the entropy $S$ (respectively) as a function of sign position in words of a given length $L$, where $L$ varies from 2 to 20. Note that the number of words of different lengths vary considerably in the data-set as shown in the side-panel in the left (showing the frequency of occurrence of words of a given length in the {\em Mieliestronk} English corpus data-base described above). Thus the variation in $G$ (or $S$) with position in the much longer words are based on statistics computed over extremely few exemplars. However, broadly we can see that the Gini index for the left-most sign is lower than that for the right-most sign in English words supporting our hypothesis. Note that, it appears that the second position from left appears to have a much higher $G$ value than either its left or right neighbour. However, this is unique to the English database and such a feature cannot be seen in other language data-bases that we have analyzed.

When we look at the variation of $S$ with sign position for words of a given length, we find that the entropy of the left-most sign is higher than the right-most sign in English words, which is again consistent with our hypothesis. The other positions do not exhibit any consistent variation except for the second sign starting from the left which (consistent with what was seen for $G$) has an entropy value that is lower than its left or right neighbour. Again, this seems to be an unique feature of the English database as we do not see it occur in the data-bases for other languages.

For example, in Figs~\ref{figs7} and \ref{figs8} we show $G$ and $S$, respectively, as a function of sign position in words of a given length $L$ for the Persian language database. As Persian is read from right to left we have numbered the rightmost sign as 1, the one left to it 2 and so on. Consistent with our hypothesis we find that $G$ for the left-most sign is higher than that for the right-most sign and the $S$ for the left-most sign is lower than that for the right-most sign. However, no other sign appears to exhibit any characteristic trend in the variation of either $G$ or $S$.
One may therefore conclude that a predictable difference in $G$ (and $S$) for the signs occurring in the left-most and right-most positions of sequences in a given language is possibly the only consistent feature seen across the languages and writing systems considered by us.
 

\pagebreak

\begin{figure*}[ht]
\includegraphics[width=0.95\linewidth]{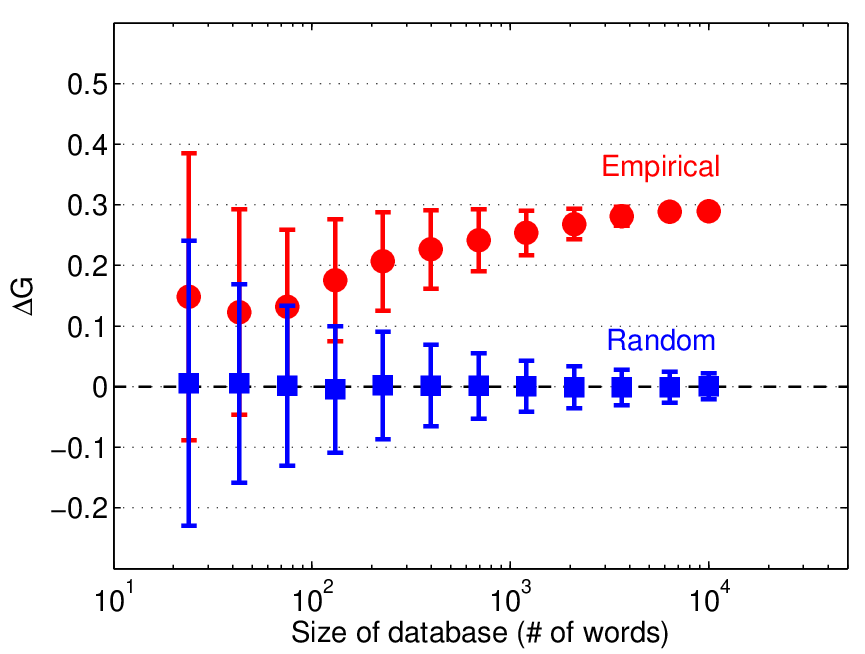}
\caption{{\bf The observed symmetry between heterogeneity of letter
occurrence probability in left and right terminal positions of words 
in a given language 
converges towards a value that is corpus size independent when the
database is sufficiently large.} 
Gini index differential $\Delta G$ shown for the left and right
terminal letter (1-gram) distributions calculated using a set of $N$
words, as a function of $N$. Empirical results are shown for random
samples (without replacement) taken from a
corpus of 10000 unique words in the Persian language,
each data point (circles) being the average over $10^3$ samples of
size $N$. For each empirical sample, a corresponding randomized sample
is created by randomly permuting the letters in each of the $N$ words,
and a data point for the randomized set (squares) represents an
average over randomizations of $10^3$ samples of size $N$. With
increasing $N$ the empirical distribution becomes distinguishable from
the randomized set (which, by definition, should not have any
left-right asymmetry) and converges to a value that is relatively 
independent of $N$. The error bars indicate standard deviation over
the different samples.
}
\label{figs3}
\end{figure*}

\begin{figure*}[ht]
\includegraphics[width=0.95\linewidth]{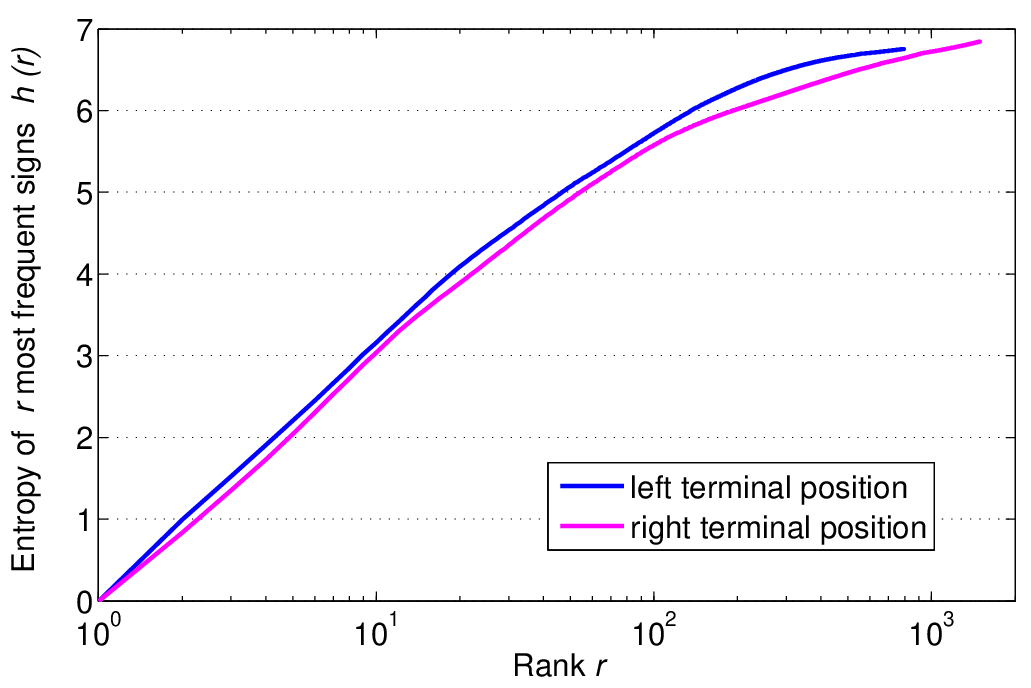}
\caption{{\bf Entropy of frequently occurring signs in the left terminal position of
 Egyptian hieroglyphic inscriptions are consistently higher than those occurring in the
 right terminal position.} However, when one considers all signs that occur at the two ends,
 the entropy difference $\Delta S$ is marginally negative, inconsistent with the
 hypothesis that in writing systems read from left to right $\Delta S$ should be positive 
 (in the Egyptian hieroglyph database used by us all sequences have been oriented so as to be 
 read from left to right). The discrepancy is explained by the fact that many more rare signs
 (i.e., appearing with extremely low frequency) occur in the right terminal position than
 at the left terminal position. This is evident from the figure, 
 where the curve corresponding to the right
 terminal position extends much further than the curve corresponding to the left terminal position.
 Note that if the entropy difference is calculated by considering the same number of signs at both ends, it is always positive which is consistent with the left-to-right orientation of the sequences according
 to the hypothesis presented here. 
}
\label{figs4}
\end{figure*}

\begin{figure*}[ht]
\includegraphics[width=0.95\linewidth]{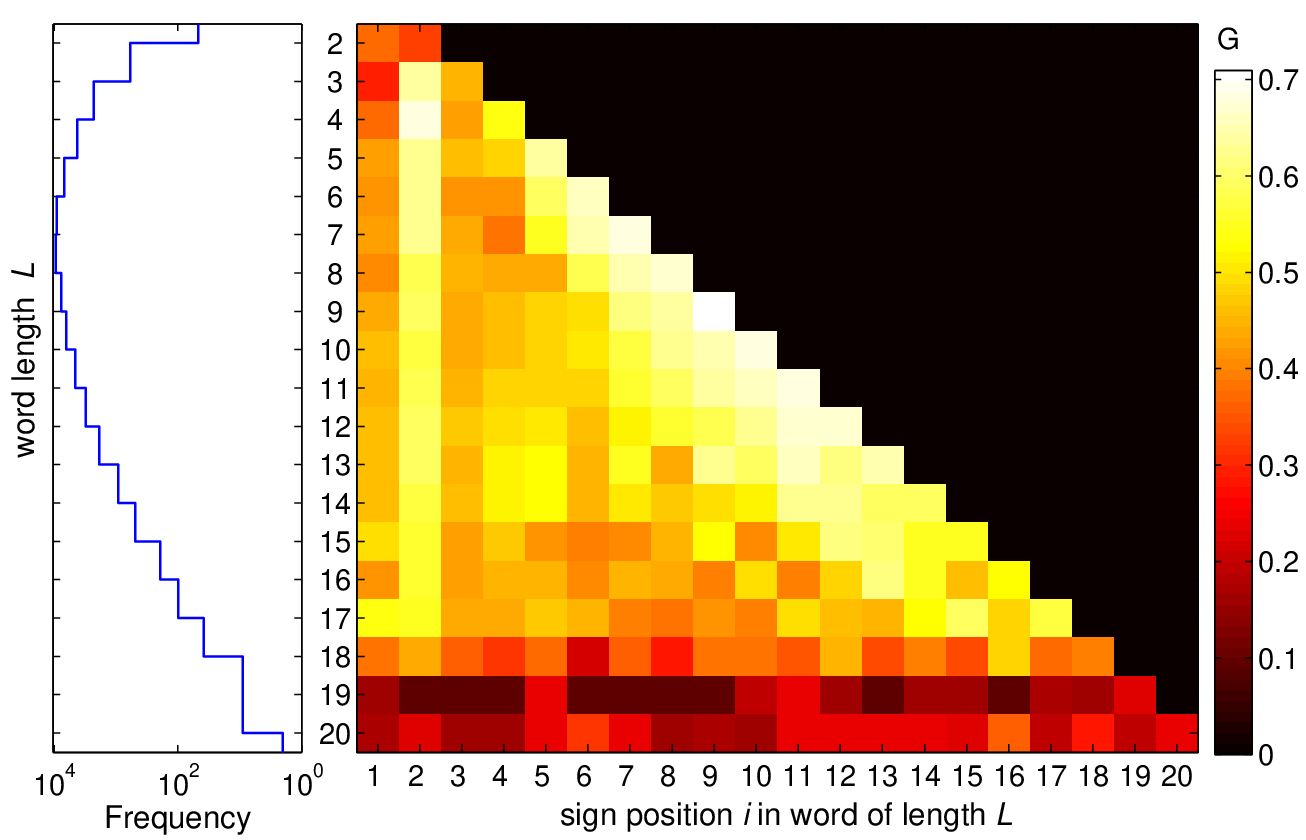}
\caption{{\bf Variation of Gini index $G$ as a function of sign position in words of a given length $L$
($2 \leq L \leq 20$) in written English.} The inequality in the distribution of occurrence probabilities for letters of the
English alphabet at different positions in words included in the {\em Mieliestronk} corpus shows
$G$ to be lower for the left-most sign compared to the right-most sign, consistent with
the direction of the writing system. As English is read from left to right, the leftmost sign is numbered as 1, the one right to it as 2, etc. The second position
(from left) appears to possess a higher value of $G$ than either end of a word, suggesting that
in English this is the most restrictive position in terms of the freedom in what signs can be used.
The left panel shows the number of words of a given length that occurs in the corpus.
Note that for longer words, the statistics is computed over very few exemplars.
The length $L$ of words is indicated along the vertical axis, while
the horizontal axis shows the position $i$ in a word of a given
length.}
\label{figs5}
\end{figure*}

\begin{figure*}[ht]
\includegraphics[width=0.95\linewidth]{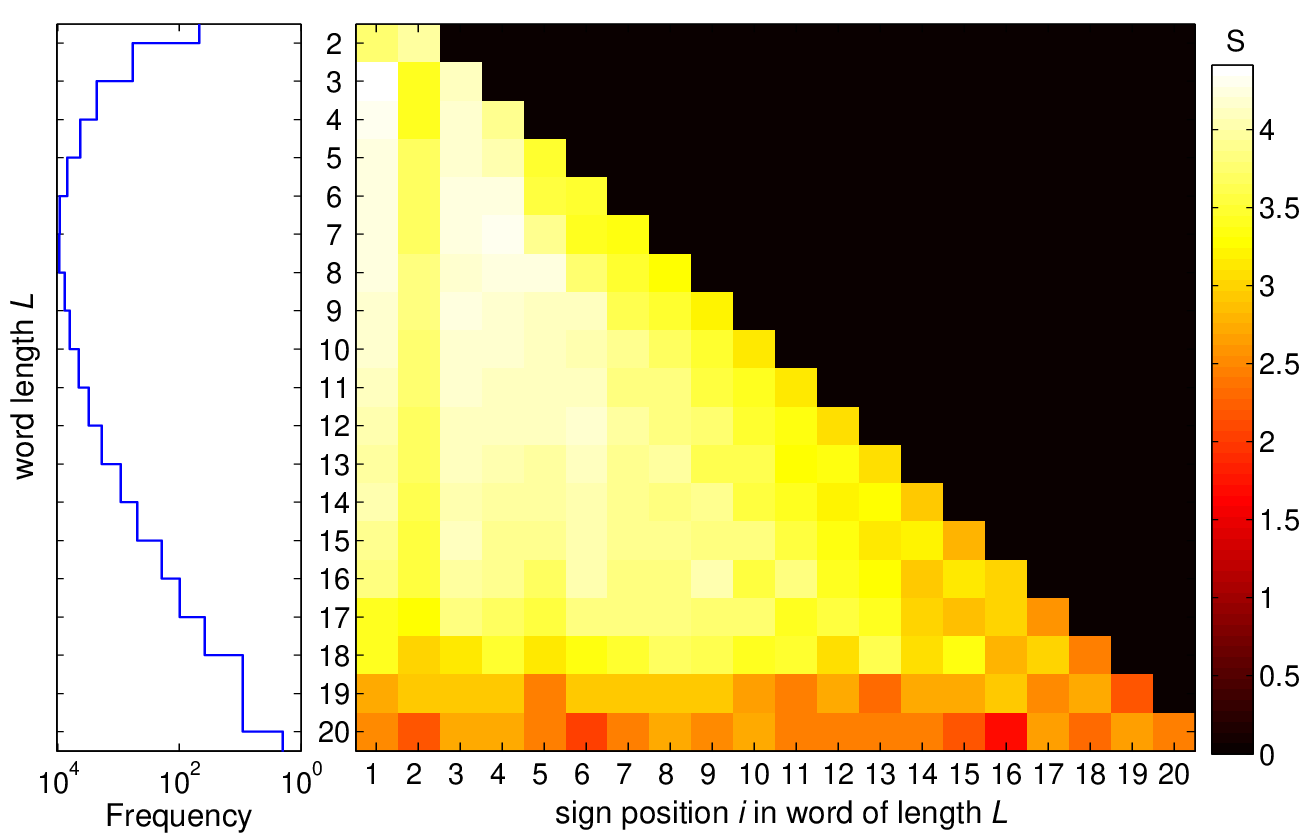}
\caption{{\bf Variation of Shannon entropy $S$ as a function of sign position in words of a given length $L$
($2 \leq L \leq 20$) in written English.} The inequality in the distribution of occurrence probabilities for letters of the
English alphabet at different positions in words included in the {\em Mieliestronk} corpus shows
$S$ to be higher for the left-most sign compared to the right-most sign, consistent with
the direction of the writing system. 
As English is read from left to right, the leftmost sign is numbered as 1, the one right to it as 2, etc. 
The second position
(from left) appears to possess a lower value of $S$ than either end of a word, suggesting that
in English this is the most restrictive position in terms of the freedom in what signs can be used
(consistent with results obtained by computing the Gini index $G$
shown in Fig~\ref{figs5}).
The left panel shows the number of words of a given length that occurs in the corpus.
Note that for longer words, the statistics is computed over very few exemplars.}
\label{figs6}
\end{figure*}

\begin{figure*}[ht]
\includegraphics[width=0.95\linewidth]{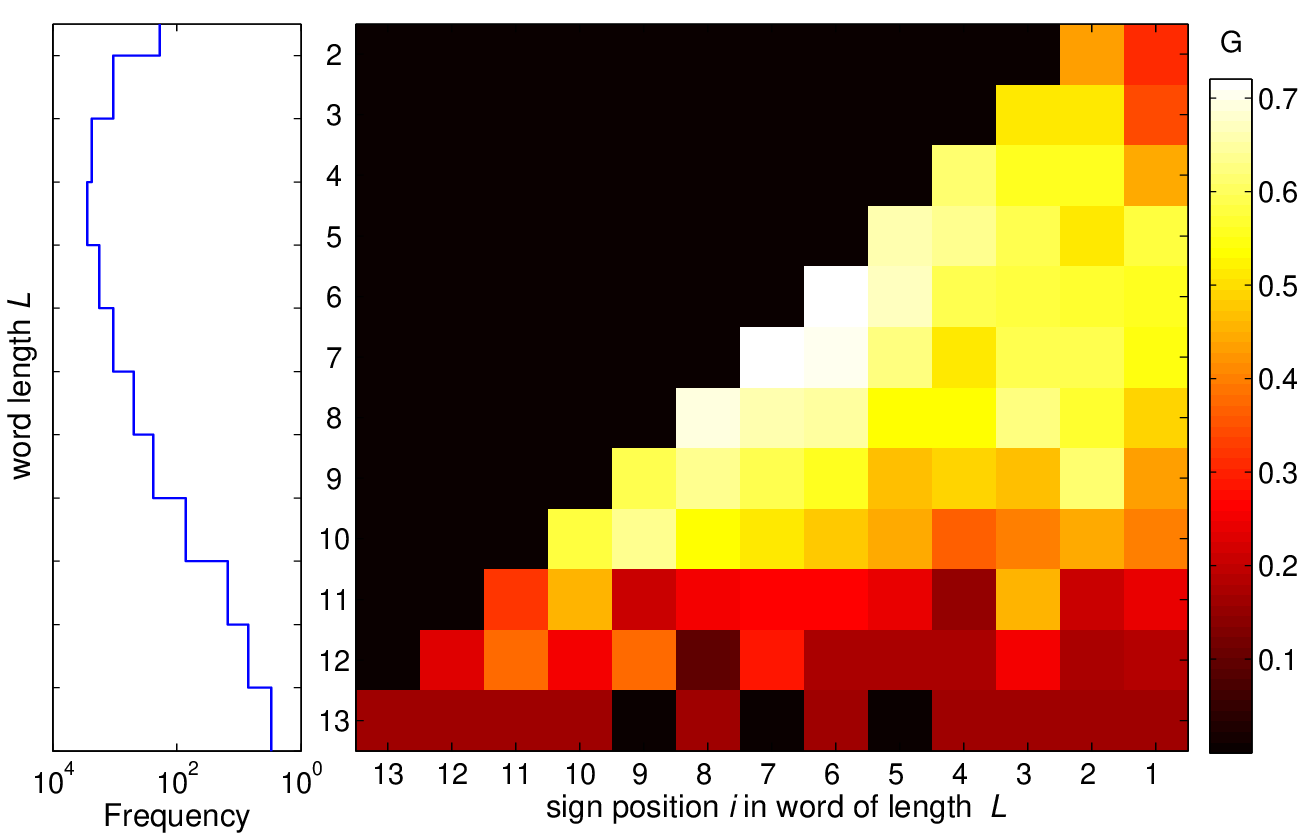}
\caption{{\bf Variation of Gini index $G$ as a function of sign position in words of a given length $L$
($2 \leq L \leq 13$) in written Persian.} The inequality in the distribution of occurrence probabilities for letters of the
Persian alphabet (a modified form of the consonantal Arabic alphabet) at different positions in words included in the Persian language corpus shows
$G$ to be lower for the right-most sign compared to the left-most sign, consistent with
the direction of the writing system.
As Persian is read from right to left, the rightmost sign is numbered as 1, the one left to it as 2, etc. 
No other position appears to have a characteristic trend in terms of sign usage inequality.
The left panel shows the number of words of a given length that occurs in the corpus.
Note that for longer words, the statistics is computed over very few exemplars.}
\label{figs7}
\end{figure*}

\begin{figure*}[ht]
\includegraphics[width=0.95\linewidth]{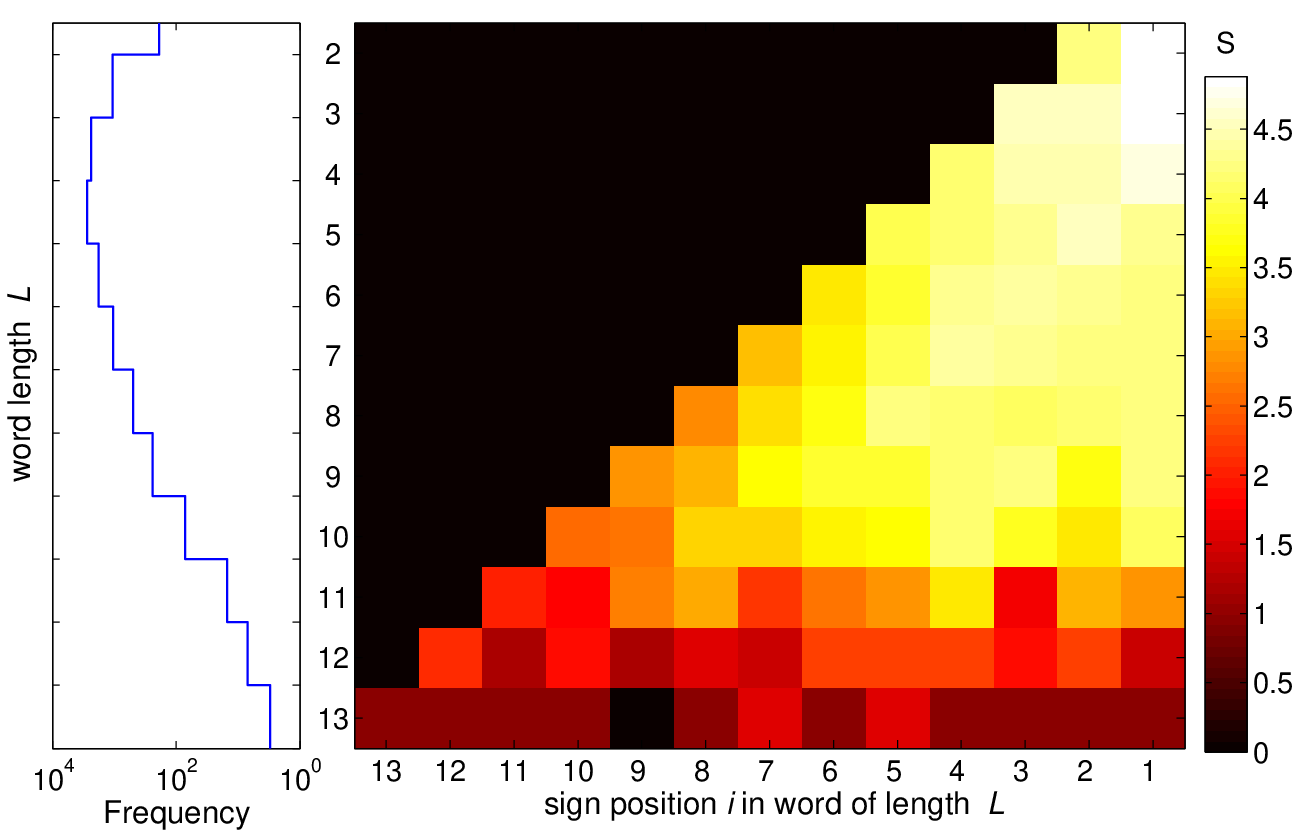}
\caption{{\bf Variation of Shannon entropy $S$ as a function of sign position in words of a given length $L$
($2 \leq L \leq 13$) in written Persian.} The inequality in the distribution of occurrence probabilities for letters of the
Persian alphabet (a modified form of the consonantal Arabic alphabet) at different positions in words included in the Persian language corpus shows
$S$ to be higher for the right-most sign compared to the left-most sign, consistent with
the direction of the writing system.
As Persian is read from right to left, the rightmost sign is numbered as 1, the one left to it as 2, etc. 
No other position appears to have a characteristic trend in terms of sign usage inequality.
The left panel shows the number of words of a given length that occurs in the corpus.
Note that for longer words, the statistics is computed over very few exemplars.}
\label{figs8}
\end{figure*}


\begin{thebibliography}{10}

\bibitem{Deacon1997}
Deacon, T.~W. {\em The symbolic species: The co-evolution of language
and the brain} (W. W. Norton, New York, N.Y., 1997)

\bibitem{Dunbar1996}
Dunbar, R. {\em Grooming, gossip, and the evolution of language}
(Faber, London, 1996).

\bibitem{Michel2011}
Michel, J.-B. {\em et al.} Quantitative analysis of culture
using millions of digitized books. {\em Science} {\bf 331,} 176-182
(2011).

\bibitem{Petersen2012}
Petersen, A. M., Tenenbaum, J. N., Havlin, S., Stanley, H. E. \& Perc,
M. Languages cool as they expand: Allometric scaling and the
decreasing need for new words. {\em Scientific Reports} {\bf 2,} 943
(2012).

\bibitem{Dodds2015}
Dodds, P. S. {\em et al.} Human language reveals a universal
positivity bias. {\em Proc. Natl. Acad. Sci. USA} {\bf 112,}
2389-2394 (2015).

\bibitem{Zipf1932}
Zipf, G. {\em Selected studies of the principle of relative
frequency in language} (Harvard University Press, Cambridge, Mass.,
1932).

\bibitem{Mitzenmacher2003}
Mitzenmacher, M. A brief history of generative models for power
law and
lognormal distributions. {\em Internet Mathematics} {\bf 1,} 226-251
(2003).

\bibitem{Cancho2003}
Ferrer i Cancho, R. \& Sol\'{e}, R. V. Least effort and the origins of
scaling in human language. {\em Proc. Natl. Acad. Sci. USA} {\bf 100,}
788-791 (2003).

\bibitem{Mumford2010}
Mumford, D. \& Desolneux, A. {\em Pattern Theory: The stochastic
analysis of
real-world signals} (A. K. Peters, Natick, Mass., 2010).

\bibitem{Shannon1948}
Shannon, C. E. A mathematical theory of communication. {\em Bell
System Tech. J.} {\bf 27,} 379-423 (1948).

\bibitem{Shannon1951}
Shannon, C. E. Prediction and entropy of printed English. {\em Bell
System Technical Journal} {\bf 30,} 50-64 (1951).

\bibitem{Schurmann1996}
Sch\"{u}rmann, T. \& Grassberger, P. The predictability of letters in
written English. {\em Fractals} {\bf 4,} 1-5 (1996).

\bibitem{Schurmann1996b}
Sch\"{u}rmann, T. \& Grassberger, P. Entropy estimation of symbol
sequences. {\em Chaos} {\bf 6,} 414-427 (1996).

\bibitem{Gini1912}
Gini C. {\em Variabilit\`{a} e Mutuabilit\`{a}: Contributo allo studio
delle distribuzioni e delle relazioni statistiche} (C. Cuppini,
Bologna, 1912) [Eng. trans. of extracts in Ceriani, L. \& Verme P. The
origins of the Gini index. {\em J. Econ. Inequal.} {\bf 10,} 421-433
(2012)].

\bibitem{Sinha2011}
Sinha, S., Chatterjee, A., Chakraborti, A. \& Chakrabarti, B. K.
{\em Econophysics: An Introduction} (Wiley-VCH, Weinheim,
2011).

\bibitem{Breiman1984}
Breiman, L., Friedman, J. H., Olshen, R. A., \& Stone, C. J.
{\em Classification and Regression Trees} (Wadsworth, Belmont, CA,
1984).

\bibitem{Plag2002}
Plag, I. {\em Word-Formation in English} (Cambridge University
Press, Cambridge, 2003).

\bibitem{Possehl2002}
Possehl, G. {\em The Indus Civilization: A Contemporary Perspective}
(AltaMira Press, Lanham, MD, 2002).

\bibitem{Sinha2011b}
Sinha, S., Ashraf, M. I., Pan, R. K. \& Wells, B. K.
Network analysis of a corpus of undeciphered Indus civilization
inscriptions indicates syntactic organization.
{\em Computer Speech and Language} {\bf 25,} 639-654 (2011).

\bibitem{Wells2015}
Wells, B. K. {\em The Archaeology and Epigraphy of Indus Writing}
(Archaeopress, Oxford, 2015).

\bibitem{Lawler2004}
Lawler, A. The Indus Script - Write or Wrong~?
{\em Science} {\bf 306,} 2026-2029 (2004).

\bibitem{Rao2009}
Rao, R., Yadav, N., Vahia, M., Joglekar, H., Adhikari, R. \&
Mahadevan, I. Entropic evidence for linguistic structure in the Indus
script. {\em Science} {\bf 324,} 1165 (2009).

\bibitem{Hunter1934}
Hunter, G. {\em The Script of Harappa and Mohenjodaro and its
connection with
other scripts} (Kegan Paul, London, 1934).

\bibitem{Mahadevan1977}
Mahadevan, I.
{\em The Indus Script: Texts, Concordance and Tables}
(Archaeological Survey of India, Calcutta, 1977).

\bibitem{Parpola1994}
Parpola, A. {\em Deciphering the Indus Script}
(Cambridge University Press, New York, 1994).

\bibitem{Daniels1996}
Daniels, P. T. \& Bright W., Eds., {\em The World's Writing
Systems} (Oxford University Press, New York, 1996).

\bibitem{Heaps1978}
Heaps, H. S. {\em Information Retrieval: Computational and Theoretical
Aspects} (Academic Press, New York, 1978).

\bibitem{McMahon2002}
McMahon, A. {\em An Introduction to English Phonology} (Edinburgh
University Press, Edinburgh, 2002).

\bibitem{DeFrancis1994}
DeFrancis, J. \& Marshall Unger, J. Rejoinder to Geoffrey Sampson.
``Chinese script and the diversity of writing systems''. {\em
Linguistics} {\bf 32,} 549-554 (1994).

\bibitem{Robinson2009}
Robinson, A. {\em Writing and Script} (Oxford University Press,
Oxford, 2009).

\bibitem{Unger2004}
Marshall Unger, J. {\em Ideogram: Chinese Characters and the Myth of
Disembodied Meaning} (University of Hawai`i Press, Honolulu, 2004).

\bibitem{Coulmas1996}
Coulmas, F. {\em The Blackwell encyclopedia of writing systems}
(Blackwell, Malden, MA, 1996).


\bibitem{Brown1994}
Brown, M. C. Using Gini-style indices to evaluate the spatial patterns of
health practitioners: Theoretical considerations and an application
based on Alberta data. {\em Social Science \& Medicine} {\bf 38,}
1243-1256 (1994).

\bibitem{Efron1987}
Efron, B. (1987) Better bootstrap confidence intervals. {\em J. Amer.
Stat. Asso.} {\bf 82,} 171-185 (1987).

\end{thebibliography}

\begin{thebibliography}{11}

\bibitem{Coulmas2003}
Coulmas, F. {\em Writing Systems: An Introduction to their Linguistic
Analysis} (Cambridge University Press, Cambridge, 2003).

\bibitem{Chao1968}
Chao, Y. R. {\em Language and Symbolic Systems} (Cambridge
University Press, Cambridge, 1968).

\bibitem{DeFrancis1984}
DeFrancis, J. {\em The Chinese Language} (University of Hawai`i Press,
Honolulu, 1984).

\bibitem{DeFrancis1989}
DeFrancis, J. {\em Visible Speech: The Diverse Oneness of Writing
Systems} (University of Hawai`i Press, Honolulu, Honolulu, 1989).

\bibitem{Sproat2000}
Sproat, R. {\em A Computational Theory of Writing Systems}
(Cambridge University Press, Cambridge, 2000).

\bibitem{Daniels1996r}
Daniels, P. T. and Bright, W., Eds. {\em The World's Writing Systems}
(Oxford University Press, New York, 1996).

\bibitem{Gardiner1957}
Gardiner, A. {\em Egyptian Grammar: Being an Introduction to the Study
of
Hieroglyphs} (3rd edn., Griffith Institute, Oxford, 1957).

\bibitem{Norvig2013}
Norvig, P., {\em English letter frequency counts:
Mayzner Revisited or ETAOIN SRHLDCU.} (2013)
Available at: http://norvig.com/mayzner.html. (Accessed: 22nd June
2015)

\bibitem{Ogden1930}
Ogden, C. K. {\em Basic English: A General Introduction with Rules
and Grammar} (Paul Treber, London, 1930). Word list available at:
http://ogden.basic-english.org/ (Accessed: 31st May 2016).

\bibitem{Kilgarriff1997}
Kilgarriff, A. Putting frequencies in the dictionary.
{\em International Journal of Lexicography} {\bf 10,} 135-155 (1997).

\bibitem{Daniels1990}
Daniels, P. T. Fundamentals of grammatology. {\em J. Amer. Oriental
Soc.} {\bf 100,} 727-731 (1990).

\bibitem{Sinha2011br}
Sinha, S., Ashraf, M. I., Pan, R. K. \& Wells, B. K.
Network analysis of a corpus of undeciphered Indus civilization
inscriptions indicates syntactic organization.
{\em Computer Speech and Language} {\bf 25,} 639-654 (2011).

\bibitem{Wells2015r}
Wells, B. K. {\em The Archaeology and Epigraphy of Indus Writing}
(Archaeopress, Oxford, 2015).

\bibitem{Mahadevan1977r}
Mahadevan, I.
{\em The Indus Script: Texts, Concordance and Tables}
(Archaeological Survey of India, Calcutta, 1977).

\bibitem{Coulmas1996r}
Coulmas, F. {\em The Blackwell Encyclopedia of Writing Systems}
(Blackwell, Malden, MA, 1996).

\bibitem{Possehl2002r}
Possehl, G. {\em The Indus Civilization: A Contemporary Perspective}
(AltaMira Press, Lanham, MD, 2002).

\bibitem{Parpola1994r}
Parpola, A. {\em Deciphering the Indus Script}
(Cambridge University Press, New York, 1994)

\bibitem{Yip2000}
Yip, P. C. {\em The Chinese Lexicon: A Comprehensive Survey} 
(Routledge, New York, 2000).
 
\bibitem{Li1981}
Li, C. N. \& Thompson, S. A. {\em Mandarin Chinese: A Functional
Reference Grammar} (University of California Press, Berkeley, 1981).

\bibitem{Bard2015}
Bard, K. A. {\em An Introduction to the Archaeology of Ancient Egypt}
(John Wiley, Chichester, 2015).

\end{thebibliography}
\end{document}